\documentclass[twoside]{article}

\usepackage[accepted]{aistats2025}
\usepackage[round]{natbib}

\usepackage{amsmath,amsthm,amsfonts,amssymb,mathtools}
\usepackage{enumitem}
\usepackage{bm,dsfont}
\usepackage[T1]{fontenc}
\usepackage{xcolor,pifont}
\usepackage{graphicx}
\usepackage{caption, subcaption, float}
\usepackage{hyperref}
\usepackage[nameinlink]{cleveref}
\usepackage[title]{appendix}
\usepackage{algorithm,algorithmic,setspace}
\usepackage{booktabs}
\usepackage{multirow}
\usepackage{makecell}
\usepackage{color, colortbl, tcolorbox}
\usepackage[hang,flushmargin]{footmisc}
\usepackage{ifthen}

\definecolor{Gray}{gray}{0.9}
\newcommand{\showcontent}{1}

\makeatletter
\def\@copyrightspace{
\@float{copyrightbox}[b]
\begin{center}
\vspace{-1.3em}
\setlength{\unitlength}{1pc}
\begin{picture}(20,2.5)
\put(0,0){\parbox[b]{19.75pc}{\small \Notice@String}}
\end{picture}
\end{center}
\end@float}
\makeatother

\newcommand{\Nc}{\mathcal{N}}
\newcommand{\Oc}{\mathcal{O}}

\newcommand{\Eb}{\mathbb{E}}
\newcommand{\Rb}{\mathbb{R}}

\newcommand{\uv}{\mathbf{u}}

\newcommand{\wv}{\mathbf{w}}
\newcommand{\xv}{\mathbf{x}}
\newcommand{\yv}{\mathbf{y}}
\newcommand{\zv}{\mathbf{z}}
\newcommand{\fv}{\mathbf{f}}

\newcommand{\mv}{\mathbf{m}}

\newcommand{\Dv}{\mathbf{D}}
\newcommand{\Hv}{\mathbf{H}}
\newcommand{\Iv}{\mathbf{I}}
\newcommand{\Rv}{\mathbf{R}}
\newcommand{\Sv}{\mathbf{S}}

\newcommand{\Xv}{\mathbf{X}}
\newcommand{\Uv}{\mathbf{U}}
\newcommand{\Zv}{\mathbf{Z}}
\newcommand{\Lv}{\mathbf{L}}
\newcommand{\Wv}{\mathbf{W}}
\newcommand{\Kv}{\mathbf{K}}

\newcommand{\argmax}{\arg\max}

\begin{document}

%

%

\runningtitle{Deep Additive Kernel}
\runningauthor{Wenyuan Zhao, Haoyuan Chen, Tie Liu, Rui Tuo, Chao Tian}

\twocolumn[

\aistatstitle{From Deep Additive Kernel Learning to Last-Layer Bayesian Neural Networks via Induced Prior Approximation
}

\aistatsauthor{ Wenyuan Zhao$^*$ \And Haoyuan Chen$^*$ \And Tie Liu}
\aistatsaddress{Texas A\&M University \\ wyzhao@tamu.edu \And Texas A\&M University \\ chenhaoyuan2018@tamu.edu \And Texas A\&M University \\ tieliu@tamu.edu}

\aistatsauthor{Rui Tuo \And Chao Tian}
\aistatsaddress{Texas A\&M University \\ ruituo@tamu.edu \And Texas A\&M University \\ chao.tian@tamu.edu}

]

\def\thefootnote{*}\footnotetext{The first two authors contributed equally.}\def\thefootnote{\arabic{footnote}}

\begin{abstract}
With the strengths of both deep learning and kernel methods like Gaussian Processes (GPs), Deep Kernel Learning (DKL) has gained considerable attention in recent years. From the computational perspective, however, DKL becomes challenging when the input dimension of the GP layer is high. To address this challenge, we propose the Deep Additive Kernel (DAK) model, which incorporates i) an additive structure for the last-layer GP; and ii) induced prior approximation for each GP unit. 
This naturally leads to a last-layer Bayesian neural network (BNN) architecture. The proposed method enjoys the interpretability of DKL as well as the computational advantages of BNN. Empirical results show that the proposed approach outperforms state-of-the-art DKL methods in both regression and classification tasks.
\end{abstract}

\section{INTRODUCTION}

Deep Neural Networks (DNNs) \citep{lecun2015deep} are powerful tools capable of capturing intricate patterns in large datasets, and have demonstrated remarkable performance across a wide range of tasks. However, DNNs are prone to overfitting on small datasets, offer limited interpretability and transparency, and lack the ability to provide uncertainty estimation. On the other hand, as a traditional kernel-based method, Gaussian processes (GPs) \citep{williams2006gaussian} are robust against overfitting. In addition, they naturally incorporate uncertainty quantification and offer enhanced interpretability and adaptability for integrating prior knowledge. 
\citet{damianou2013deep} proposed Deep Gaussian Processes (DGPs) by stacking multiple layers of GPs, which introduces the hierarchical structure of deep learning with the probabilistic, non-parametric nature of GPs. Although the deep structure of DGPs allows them to learn features at different levels of abstraction, the tuning and optimization of DGPs, particularly for large datasets, can be difficult due to the layered structure, requiring careful consideration of hyperparameters and approximation techniques.

To learn rich hierarchical representations from the data with proper interpretability and uncertainty estimation, \citet{wilson2016deep} introduced Deep Kernel Learning (DKL) by incorporating DNN into the last layer kernel methods. This hybrid model enhances both flexibility and scalability compared to pure GPs, making it well-suited for a variety of real-world tasks, including regression, classification, and active learning, especially in scenarios where uncertainty quantification is critical.


One of the main challenges in DKL arises from the GP layer, which requires $\Oc(N^3)$ training time for $N$ data points, limiting its scalability for large datasets. Although DKL scales better than pure GPs as the feature dimension is reduced by the NN encoder, computing DKL exactly is still expensive, when the extracted features are high-dimensional, particularly in image (or video) datasets. To address this issue, several methods have been developed for GP approximation, including Random Fourier Features (RFF) \citep{rahimi2007random}, Stochastic Variational GP (SVGP) \citep{titsias2009variational,hensman2015scalable}, and Kernel Interpolation for Scalable Structured GP (KISS-GP) \citep{wilson2015kernel}, and these have been incorporated into DKL models \citep{wilson2016stochastic,xue2019deep,xie2019deep}. However, these sparse GP approximations via inducing points require $\mathcal{O}(M^3)$ time to compute the Evidence Lower Bound (ELBO), where $M$ is the number of inducing points. Therefore, DKL remains inefficient when a large number of inducing points are necessary for complex ML tasks.

Another challenge in DKL, as identified by \citet{ober2021promises}, is the tendency to overcorrelate features to minimize the complexity penalty term in the marginal likelihood. A fully Bayesian method incorporating Markov Chain Monte Carlo (MCMC) can effectively resolve this problem, but exhibit poor scalability with high-dimensional posterior distributions. In response, \citet{matias2024amortized} proposed Amortized Variational DKL (AV-DKL), which uses NN-based amortization networks to determine the inducing locations and variational parameters through input-dependent sparse GPs \citep{jafrasteh2021input}, thereby attenuating the overcorrelation of NN output features. 


In this work, we propose the \textbf{D}eep \textbf{A}dditive \textbf{K}ernel (DAK) model, which embeds hierarchical features learned from NNs into additive GPs, and interpret the last-layer GP as a Bayesian neural network (BNN) layer \citep{mackay1992practical,harrison2024variational} with a sparse kernel activation via induced prior approximation on designed grids \citep{ding2024sparse}. The proposed methodology jointly trains the variational parameters of the last layer and the deterministic parameters of the feature extractor by maximizing the variational lower bound. This hybrid architecture enjoys the mathematical interpretability of DKL as well as the computational advantages of BNN, and 
also leads to a closed-form ELBO and predictive distribution for regression tasks, bypassing Monte Carlo (MC) sampling during inference and training. Our contributions are as follows:
\vspace{-0.3cm}
\begin{itemize}[itemsep=1.5pt,,parsep=1.5pt]
    \item We introduce a DAK model that reinterprets the deep additive kernel learning as a last-layer BNN via induced prior approximation. The proposed DAK enjoys the mathematical interpretability of DKL as well as the computational advantages of BNN, and can be adapted to a variety of DL applications.
    \item We derive closed-form expressions of both the predictive distribution in inference and the ELBO during training for regression tasks, eliminating the need for sampling. The reduced computational complexity is linear to the size of the induced grid.
    \item In experimental studies, we demonstrate that the proposed DAK model outperforms state-of-the-art DKL methods in both regression and classification tasks, while also mitigating the overfitting issue. Source code of DAK is available at the following link \url{https://github.com/warrenzha/dak2bnn}.
\end{itemize}
\vspace{-0.2cm}
The remainder of this paper is organized as follows: \Cref{sec:prelim} introduces the background on GPs, additive GPs, and DKL. In \Cref{sec:dak}, we present the proposed DAK model and discuss its computational complexity compared to other state-of-the-art DKL models, followed by related work in \Cref{sec:related work}. We present the experimental results in \Cref{sec:exp}, and conclude the paper in \Cref{sec:conc}.

\section{PRELIMINARIES}
\label{sec:prelim}
\paragraph{GPs.}
A GP $f(\cdot) \sim \mathcal{GP}(\mu(\cdot) ,k(\cdot,\cdot) )$ is completely specified by its mean function $\mu(\cdot)$ and the covariance (kernel) function $k(\cdot,\cdot)$. Given a dataset $\mathcal{D}=\{ \Xv, \yv \}$, where $\Xv=\{ \xv_{i} \in \mathbb{R}^D \}_{i=1}^{N}$ are training points and $\yv= ( y_1,\ldots,y_N )^{\top}$ is the corresponding observation where $y_i\in \Rb$, the standard procedure of GPs assumes $\mu(\cdot)$ is a constant function assigned to zero and considers $y_i=f(\xv_i) + \epsilon_i$ with Gaussian noise $\epsilon_i \sim \mathcal{N}(0, \sigma^2_f)$ for $i=1,\ldots,N$. The predictive posterior $\fv_{\ast}:=f(\Xv^{\ast})$ 
at $N_{\ast}$ test points $\Xv^{\ast}:= \{ \xv_{i}^{\ast} \in \mathbb{R}^D \}_{i=1}^{N_{\ast}}$ can also be expressed in closed form as a Gaussian distribution:
\begin{align}
    & \fv_{\ast} \vert \Xv,\yv,\Xv^{\ast} \sim \mathcal{N} \left( \bm{\mu}_{\ast} , \bm{\Sigma}_{\ast} \right), \\
    \bm{\mu}_{\ast} &= \mathbf{K}_{\Xv^*,\Xv} \left( \mathbf{K}_{\Xv,\Xv} + \sigma_f^2 \Iv \right) ^{-1} \yv,\label{eq:GPR mean}\\
    \bm{\Sigma}_{\ast} &= \mathbf{K}_{\Xv^*,\Xv^*}
    - \mathbf{K}_{\Xv^*,\Xv} \left( \mathbf{K}_{\Xv,\Xv} + \sigma_f^2 \Iv \right)^{-1} \mathbf{K}_{\Xv,\Xv^*},
\end{align}
where $\mathbf{K}_{\Xv,\Xv'}$ denotes the kernel matrix with $[\mathbf{K}_{\Xv,\Xv'}]_{ij}:=k(\xv_{i}, \xv'_{j})$. 
The common challenges of GPs are the $\mathcal{O}(N^3)$ computational complexity of inference and the ``curse of dimensionality'' with high-dimensional data. We refer the readers to \citet{williams2006gaussian} for more details. 

\paragraph{Additive GPs.}
\citet{duvenaud2011additive} introduced the additive GP model, which allows additive interactions of all orders, ranging from first-order interactions all the way to $D$-th order interactions. In this work, we consider the simplest case where the additive GP is restricted to the first order with the same base kernel for each unit $d \in \{1,\ldots,D\}$:
\begin{align}
\label{eq:additiveGP}
    f(\xv)=\sigma\sum_{d=1}^{D} g_d (x_d)+\mu(\xv),
\end{align}
where $x_d$ is the $d$-th feature of the point $\xv\in \Rb^D$, $g_d(x_d) \sim \mathcal{GP}(0,k_d(x_d,x_d^{\prime}))$ is the \emph{centered} base GP unit. The resulting $f(\xv): \Rb^D\rightarrow \Rb$ is a GP specified by mean function $\mu(\xv)$ and additive kernel $k^{[D]}_{\text{add}}(\xv, \xv^{\prime}) = \sigma^2 \sum_{d=1}^{D}k_d(x_d, x_d^{\prime})$, where $\sigma^2$ is the variance assigned to all first-order interactions. \citet{delbridge2020randomly} showed that additive kernels projecting to a low dimensional setting can match or even surpass the performance of kernels operating in the original space. However, an additive kernel alone offers no computational advantage, as the cubic complexity remains. As we will show shortly, the additive GP perspective allows us to apply an induced prior approximation, which reduces computational costs and leads to a last-layer Bayesian representation.

\paragraph{Deep Kernel Learning.} 
The performance of GPs is limited by the choice of kernel. DKL \citep{wilson2016deep} attempts to solve this problem by DNNs, of which the structural properties can model high-dimensional data and large datasets well. DKL first learns feature transformations $h_\psi(\cdot)$ of DNNs with the parameters $\psi$. The outputs of DNNs are then used as inputs to a GP layer $\mathcal{GP}\left( \mu_{\theta}(\cdot ),k_{\theta}(\cdot ,\cdot ) \right)$ with the parameters $\theta$ resulting in the effective kernel $k_{\text{DKL}}\left( \xv,\xv^{\prime} \right) =k_{\theta}\left( h_{\psi}(\xv), h_{\psi}(\xv^{\prime}) \right)$ to learn uncertainty representation provided by GPs. The complete set of parameters $\left\{ \psi ,\theta \right\}$ in DNNs and GPs are trained jointly by maximizing the marginal log-likelihood (MLL).

\section{DAK: Deep Additive Kernel}
\label{sec:dak}
In GPs, the kernel function defines the similarity between data points, playing a crucial role in the model's predictions. However, choosing the right kernel for complex, high-dimensional data can be challenging. DKL addresses this by learning the kernel function directly from data using a DNN. In practice, the outputs of DNNs can still be too complex to scale GPs, e.g. multi-task regression, or image recognition. Therefore, we present the Deep Additive Kernel as a last-layer BNN that can efficiently reduce GPs to single-dimensional ones without loss in performance.

\paragraph{Model.} 
We present the construction of the DAK using the proposed additive GPs. 
Let $h_{\psi}: \Rb^D \rightarrow \Rb^{P}$ be a neural network, and we consider a total of $P$ \emph{centered} one-dimensional base GPs $g_p \sim \mathcal{GP}(0, k_p(\cdot, \cdot))$ with Laplace kernel $k_p(x_p, x_p^{\prime})= \exp\left( - \vert x_p - x_p^{\prime} \vert / \theta_p \right)$ 
for $p=1,\ldots,P$. The forward pass of deep additive model $f(\cdot): \Rb^D \rightarrow \Rb$ with Laplace kernel for each base GP 
at $N$ data points $\Xv$ is described as follows:
\begin{align}
    f (\Xv) &=\sum_{p=1}^{P} \sigma_{p}g_{p} \Big( 
    \begingroup
    \color{black}
        \underbrace{ 
            \color{black} 
            h^{[p]}_{\psi}\left( \Xv \right) 
        }_{ \color{black} := \Hv_p \in \Rb^{N} }
    \endgroup
    \Big) + \mu \label{eq:deepadditive},
\end{align}
where $\Hv_p:=h^{[p]}_{\psi}\left( \Xv \right) \in \Rb^{ N}$ is the $p$-th feature vector of neural network representations $h_{\psi}\left( \Xv \right) \in \Rb^{N \times P}$, and $\mu \in \mathbb{R}$ is a constant prior mean placed on $f(\cdot)$. The resulting deep additive kernel can be written as:
\begin{align}
    k_{\text{DAK}}\left( \xv,\xv^{\prime} \right) 
    &= k^{[P]}_{\text{add}}\left( h_{\psi}(\xv),h_{\psi}(\xv^{\prime}) \right) \nonumber \\
    &= \sum_{p=1}^{P} \sigma_{p}^{2} k_{p}\Big(
    \begingroup
        \color{black}
        \underbracket{
            \color{black} h_{\psi}^{[p]}(\xv)
        }_{ \color{black} h_{p}\in \Rb
        }
    \endgroup
    , 
    \begingroup
        \color{black}
        \underbracket{
            \color{black} h_{\psi}^{[p]}(\xv^{\prime})
        }_{ \color{black} h_{p}^{\prime}\in \Rb
        }
    \endgroup
    \Big).
\end{align}
The corresponding coefficients $\{\sigma_{p} \}_{p=1}^{P}$ help learn a correlated contribution of each base kernel, and high-dimensional features of neural network $h_{\psi}\left( \Xv \right)$ can be solved in a single-dimensional space $\Rb$ to achieve scalability. The GP hyperparameters can be inherently optimized with DNN parameters $\psi$ without additional computational cost in our method. The details of reparameterization and optimization tricks are deferred when we discuss \textbf{Inference} and \textbf{Training} shortly.

\paragraph{Induced Prior Approximation.}
An essential ingredient of the proposed method is to apply a reduced-rank approximation to the prior, i.e., the base GPs. This approximation is referred to as the induced prior approximation \citep{ding2024sparse}, given by
\begin{align}
    \tilde{g}_{p} (\cdot ) &:=\Kv_{(\cdot),\Uv} \Kv_{\Uv,\Uv}^{-1} \, g_{p}(\Uv) \label{eq:inducedAppro}\\
    &=
    \begingroup
    \color{black}
        \underbrace{ 
            \color{black}
            \Kv_{(\cdot),\Uv} \left[ \Lv_{\Uv}^{\top} \right]^{-1}
        }_{\color{black}
            := \phi(\cdot) \in \Rb^{1\times M}
        }
    \endgroup
    \begingroup
    \color{black}
        \underbrace{ 
            \color{black}
            \Lv_{\Uv}^{-1} g_{p} ( \Uv ) 
        }_{\color{black}
        :=\zv_p \,\sim\, \mathcal{N}(\bm{0}, \bm{I}_{M})
        }
    \endgroup  \nonumber\\
    &=\phi \left( \cdot \right) \zv_p, \label{eq:GPlayer}
\end{align}
where $\Uv=\{ u_{i}\in \mathbb{R} \}_{i=1}^{M}$ denotes the induced grids (see details in \Cref{sec:sparse chol decompose}), $\Lv_\Uv \in \Rb^{M\times M}$ is a lower triangular matrix derived by the Cholesky decomposition of $\Kv_{\Uv,\Uv}=\Lv_{\Uv} \Lv^{\top}_{\Uv}$, and $\zv_p=\Lv_{\Uv}^{-1} g_{p} \left( \Uv \right)\sim \mathcal{N}(\mathbf{0},\bm{I}_M)$ are i.i.d. standard Normal random variables. The main idea of induced prior approximation (\cref{eq:inducedAppro}) is to use the GP regression (see \cref{eq:GPR mean}) to reconstruct the prior GP. According to the theory of GP regression \citep{yakowitz1985comparison,wang2020prediction}, $\tilde{g}_{p}$ converges to the original prior as $\Uv$ becomes dense in its domain.

The standard Cholesky decomposition of a dense matrix requires $O(M^3)$ time. However, by leveraging the Markov property of the Laplace kernel, the decomposed matrix $\Lv_{\Uv}^{-1}$ becomes sparse if $\Uv$ is designed by a one-dimensional dyadic point set with increasing order \citep{ding2024sparse}. As a consequence, the complexity of Cholesky decomposition can be reduced to $\Oc(M)$. Details of obtaining the sparse Cholesky factors are presented in \Cref{sec:sparse chol decompose}. 

Applying the sparsely induced GP approximation in \cref{eq:GPlayer} together with the deep additive model in \cref{eq:deepadditive}, we obtain the final DAK:
\begin{align}
    \tilde{f}(\Xv) &=\sum_{p=1}^{P} \sigma_p \tilde{g}_{p} \left( h^{[p]}_{\psi}(\Xv) \right) + \mu \notag \\
    &=\sum_{p=1}^{P} \sigma_{p} \Big(
    \phi \big(\Hv_p \big) \zv_{p}\Big) + \mu,  \label{eq:additiveDKL} 
\end{align}
where $\phi(\Hv_p):=\Kv_{\Hv_p,\Uv} \left[ \Lv_{\Uv}^{\top} \right]^{-1}\in \Rb^{1\times M}$ denotes the kernel activation of $p$-th base GP, $\mu$ is the mean of additive GP, and $\zv_p$'s are random weights with  i.i.d. normal prior distribution $\zv_p \sim \Nc(\bm{0},\bm{I}_M)$, for all $p=1,\ldots,P$.

\begin{figure*}[ht]
    \centering
    \includegraphics[width=0.95\textwidth]{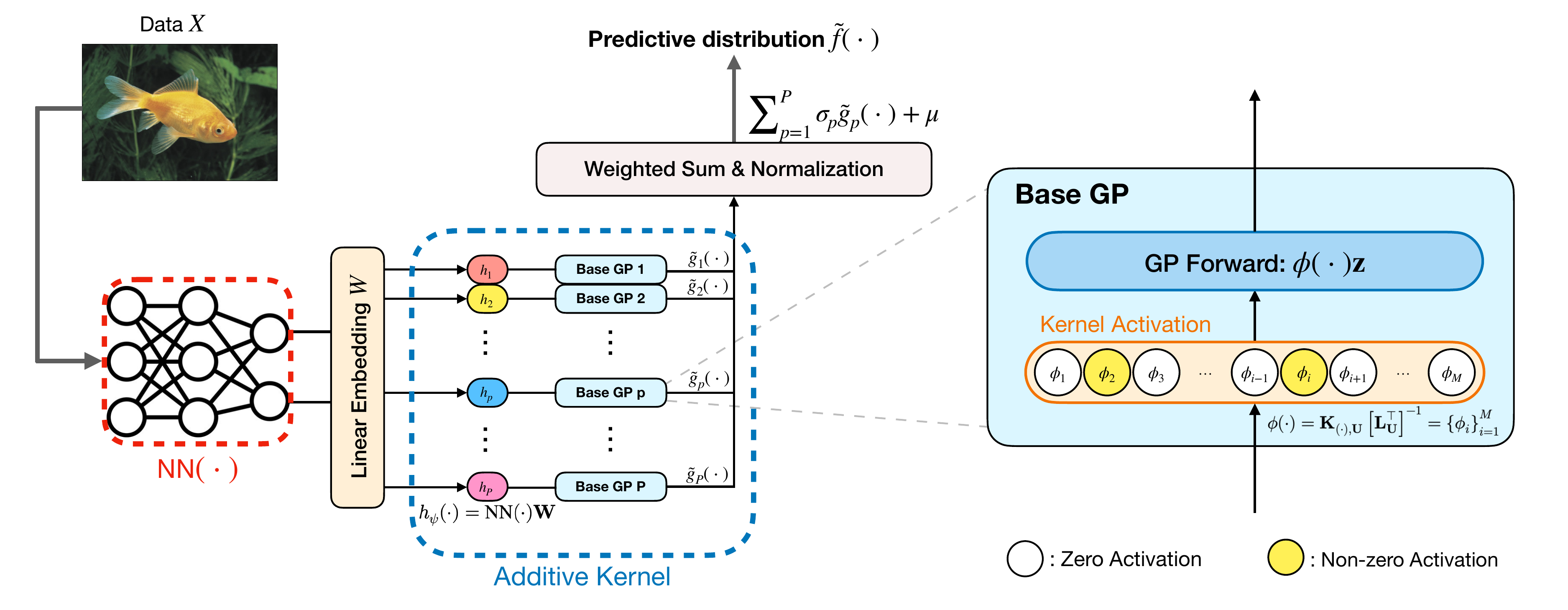}
    \caption{\small{Model architecture of Deep Additive Kernel (DAK). DAK consists of a feature extractor $\text{NN}(\cdot)$ with a linear embedding layer $\Wv$, an additive kernel with base GP $\tilde{g}_{p}(\cdot)$ for $p=1,\ldots,P$, and a weighted sum layer. The embedded features learned by DNN are decomposed as first-order components and fed to base GPs, each consisting of a kernel activation and a GP forward layer. Each kernel activation is designed by a one-dimensional dyadic point set on an induced grid with sparse non-zero activated neurons.}}
    \label{fig:model}
\end{figure*}

The proposed DAK in \cref{eq:additiveDKL} results in deep additive kernel learning which mathematically possesses the form of a last-layer BNN \citep{mackay1992practical} but with a kernel activation $\phi(\cdot):=\Kv_{(\cdot),\Uv} \left[ \Lv_{\Uv}^{\top} \right]^{-1}$ followed by a Gaussian forward layer with i.i.d. prior weights under standard Normal distribution. 

\Cref{fig:model} illustrates the module architecture of DAK. Thanks to the mathematical equivalence to BNN and reparameterization tricks on learnable parameters, all modules of our model can be implemented either as a deterministic or a Bayesian NN layer. This results in a single NN with hybrid layers, which allows us 
to straightforwardly take advantage of parallelized GPU computing using the readily available PyTorch package \citep{paszke2019pytorch}.

\paragraph{Inference.}
To obtain the predictive distribution, we perform Variational Inference (VI) to estimate the posterior using DAK in \cref{eq:additiveDKL}. With a motivation to naturally interpret the deep additive kernel learning as a last-layer BNN, and taking the mean field assumptions, we select a variational family of independent but not identical Gaussian weights $\Zv:=[ \zv_1,\ldots,\zv_P ]$ and additive Gaussian bias $\mu$, denoted by $\Theta_{\text{var}}:=\left\{ \{ \zv_{p}\}_{p=1}^{P}, \mu \right\}$. The variational Gaussian weights are parameterized as $\zv_{p} \sim \mathcal{N} (\bm{m}_{\zv_p} ,\Sv_{\zv_p})$ for $p=1,\ldots,P$, and the variational bias is parameterized as $\mu \sim \mathcal{N} ( m_{\mu},\sigma^2_{\mu} )$. We denote the reparameterization of $\Theta_{\text{var}}$ as $\bm{\eta}:=\left\{ \{ \mv_{\zv_{p}},\Sv_{\zv_{p}}\}_{p=1}^{P} , \{m_{\mu},\sigma_{\mu}\} \right\}$. 

Note that $\Sv_{\zv_p}\in\Rb^{M \times M}$ is a diagonal covariance matrix due to the mean field assumption of $\zv_p$, where $M$ is decided by the size of induced interpolation grids defined in \cref{eq:GPlayer}. The variational distribution is given by $q_{\bm{\eta}}(\Theta_{\text{var}}) = q(\mu) \prod_{p=1}^{P} q(\zv_{p}) = \Nc ( m_{\mu} ,\sigma_{\mu}^2 )\prod_{p=1}^{P} 
\Nc ( \bm{m}_{\zv_p} ,\Sv_{\zv_p} )$, and the prior of $\Theta_{\text{var}}$ is denoted by $p(\Theta_{\text{var}})$.

The other deterministic parameters consist of DNN parameters $\psi$ and additive GP hyperparameters $\left\{ \bm{\sigma} \right\}:=[\sigma_1,\ldots,\sigma_P]^{\top}$, denoted as $\bm{\theta}:=\{\psi, \bm{\sigma} \}$. We treat GP lengthscales as fixed parameters specified during initialization and apply an additional linearly embedding layer to DNN: $h_{\psi}(\cdot ):=\text{NN} (\cdot )\Wv: \Rb^D \rightarrow \Rb^P$, where $\text{NN} (\cdot ): \Rb^D \rightarrow \Rb^{D_w}$ represents any DNN, and $\Wv\in \Rb^{D_{w}\times P}$ is the linear embedding. The lengthscales can be inherently optimized by learning the NN weights $\Wv$ without loss in performance, which is encoded in the DNN parameters $\psi$. Further details are provided in \Cref{sec:theo}. 

Given a data point $\xv\in \Rb^D$, the forward pass of predictive distribution is given by
\begin{align}
\label{eq:DAK prediction}
    \tilde{f}_{\xv}:= \tilde{f}(\xv; \bm{\theta}, \bm{\eta}) =\sum_{p=1}^{P} \sigma_{p} \Big(
    \phi \big(h^{[p]}_{\psi}(\xv) \big) \zv_{p}\Big) + \mu,
\end{align}
where the complete set of parameters is $\Theta:= \left\{ \bm{\theta}, \bm{\eta} \right\} = \left\{ \psi ,\bm{\sigma}, \{ \mv_{\zv_{p}},\Sv_{\zv_{p}}\}_{p=1}^{P} , \{m_{\mu},\sigma_{\mu}\} \right\}$, consisting of the deterministic parameters $\bm{\theta} := \left\{ \psi ,\bm{\sigma} \right\}$ and the variational parameters $\bm{\eta}:= \left\{ \{ \mv_{\zv_{p}},\Sv_{\zv_{p}}\}_{p=1}^{P} , \{m_{\mu},\sigma_{\mu}\} \right\}$. In \Cref{sec:uq of inference}, we derive an analytical form for the predictive distribution $\tilde{f}$.



\paragraph{Training.} 
Vanilla DKL optimizes the marginal log-likelihood $\log \text{Pr} \left( \yv \vert \Xv, \bm{\theta} \right)$ which involves intractable integral of non-conjugate likelihoods in some tasks such as classification. We apply the framework of stochastic variational inference to fit GPs via the variational distribution $q_{\bm{\eta}}(\Theta_\text{var})$. Consequently, during training, we optimize the variational lower bound, as formulated in \cite{hensman2015scalable}, to achieve efficient and scalable inference:
\begin{align}
\label{eq:VI lower bound}
\log \text{Pr}(\yv | \Xv, \bm{\theta}) \geq &\sum_{\xv, y \in \Xv, \yv}\Eb_{q_{\bm{\eta}}(\Theta_{\text{var}})} \left[ \log \text{Pr} \big(y | \tilde{f}_{\xv} \big) \right] \nonumber \\
& - \text{KL} \left[ q_{\bm{\eta}}(\Theta_{\text{var}} ) \| p(\Theta_{\text{var}}) \right].
\end{align}
The details of the training are derived in \Cref{sec:training}. The resulting objective is known as Evidence Lower Bound (ELBO):
\begin{align}\label{eq:elbo}
    \mathcal{L} (\bm{\theta}, \bm{\eta}) :=
    & \  {\Eb}_{q_{\bm{\eta}}(\Theta_{\text{var}} )} \left[ \log \text{Pr} (\yv \vert \tilde{f}_{\Xv}) \right] \nonumber \\
    & - \text{KL} \left[ q_{\bm{\eta}}(\Theta_{\text{var}} ) \| p(\Theta_{\text{var}}) \right].
\end{align}
The first term of ELBO, the expected log-likelihood, can be estimated by MC methods. To avoid the potential computing cost of sampling, we also derive a sampling-free analytical ELBO with a closed form for regression tasks. The details of the derivation are provided in \Cref{sec:elbo}.

\paragraph{Computational Complexity.}
We summarize the computational complexity of our proposed DAK model compared to other state-of-the-art GP and DKL methods in \Cref{tab:complexity}. On the one hand, the number of inducing points $\hat{M}$ in SVGP and KISS-GP may need to be chosen quite large in more complex or multitask GPs since it is associated with the GP input dimensionality, while at competitive performances the size of induced grids $M$ in DAK can be small due to the first-order additive structure. On the other hand, the dimension of the embedding layer $P$ is usually smaller than the dimension of NN outputs $D_w$. For tasks that require MC sampling, a small number of samples is usually sufficient due to the sampling efficiency of the BNNs. Further discussion is deferred to \Cref{sec:complexity}.

\begin{table}[tb!]
    \caption{\small{Computational complexity of DKL models for $N$ training points in one iteration. $\hat{M}$ is the number of inducing points in SVGP and KISS-GP, while $M$ is the size of induced grids in DAK, $M < \hat{M}$. $S$ is the number of MC samples, $B$ is the size of mini-batch, $D_w$ is the dimension of the NN outputs in DKL, $P$ is the dimension after the linear embedding of NN features. DAK-MC refers to DAK using MC approximation, while DAK-CF refers to DAK using closed-form inference and ELBO.}}
    \centering
    \resizebox{\columnwidth}{!}{
    \begin{tabular}{lcc}
    \toprule[1pt]
                  & \textbf{Inference}       & \textbf{Training} (per iteration) \\
    \midrule[0.5pt]
    NN + SVGP     & $\Oc(\hat{M}^2 N)$    & $\Oc( S D_w \hat{M}B + \hat{M}^3)$ \\
    NN + KISS-GP  & $\Oc(D_w \hat{M}^{1+\frac{1}{D_w}})$  & $\Oc(S D_w \hat{M}B + D_w \hat{M}^{\frac{3}{D_w}})$ \\
    DAK-MC (ours) & $\Oc(SM)$       & $\Oc(SPMB + PM)$   \\
    DAK-CF (ours) & $\Oc(M)$        & $\Oc(PMB + PM)$    \\
    \bottomrule[1pt]
    \end{tabular}
    }
    \label{tab:complexity}
\end{table}

\paragraph{Remarks.}
We highlight several key aspects of the proposed model: 
\textbf{1)} The proposed model adds base GP components directly, rather than adding kernels as described in \citep{duvenaud2011additive}. While they are mathematically equivalent, using an additive kernel alone does not lead to any computational advantage over any standard kernel, where the cubic computational complexity persists. In contrast, the additive GP led us to apply the induced approximation technique, which naturally leads to efficient computation and the last-layer Bayesian interpretation. 
\textbf{2)} The induced prior approximation in our model differs from the standard inducing points approximation \citep{titsias2009variational} often used for GPs. In our approach, we approximate the prior using a fixed set of induced grids, resulting in a BNN representation. In contrast, the standard inducing point methods treat the inducing points as variational parameters for optimization, which does not lead to a BNN representation. Our method is much easier to implement and has a theoretical guarantee that as the inducing locations become dense in the input region, the approximation will become exact.
\textbf{3)} In the forward pass $f(\cdot)$ as defined in \cref{eq:DAK prediction}, we chose to place a constant prior to the mean $\mu$ on $f(\cdot)$, which also naturally facilitates the construction of the last-layer BNN. This provides a bridge between canonical GPs and BNNs, making them more flexible and extendable to diverse applications.

\begin{figure*}[ht]
    \centering
    \subfloat[\small{Exact GP.} \label{fig:gp1d}]{\includegraphics[width=.2\textwidth]{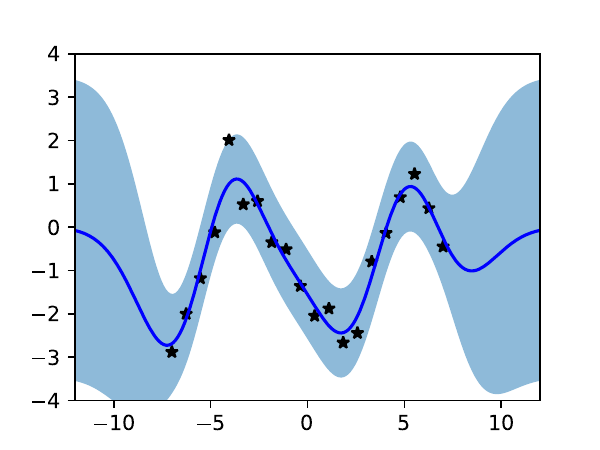}}
    \subfloat[\small{DGP.} \label{fig:dgp1d}]{\includegraphics[width=.2\textwidth]{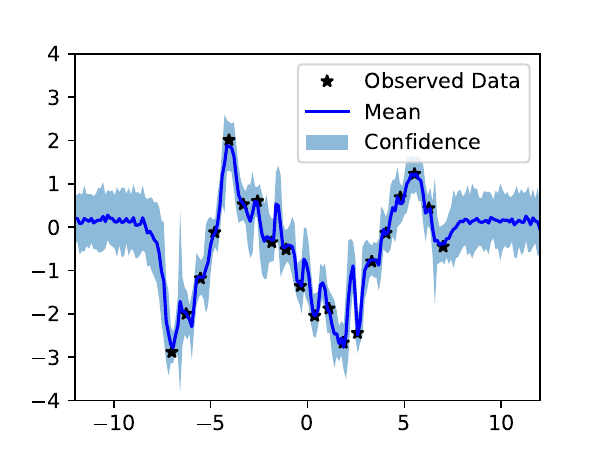}}
    \subfloat[\small{Exact DKL.} \label{fig:dkl1d}]{\includegraphics[width=.2\textwidth]{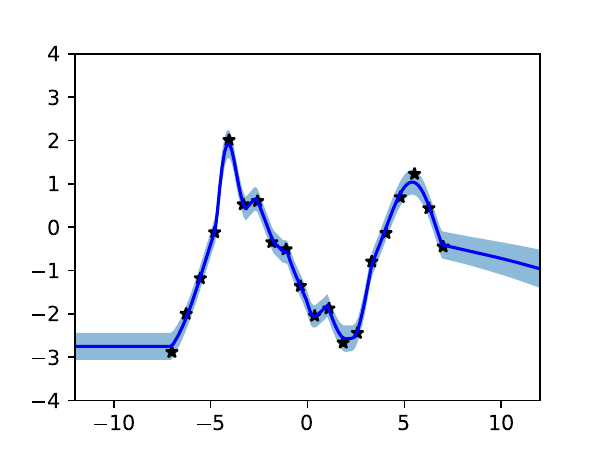}}
    \subfloat[\small{DAK.} \label{fig:dak1d}]{\includegraphics[width=.2\textwidth]{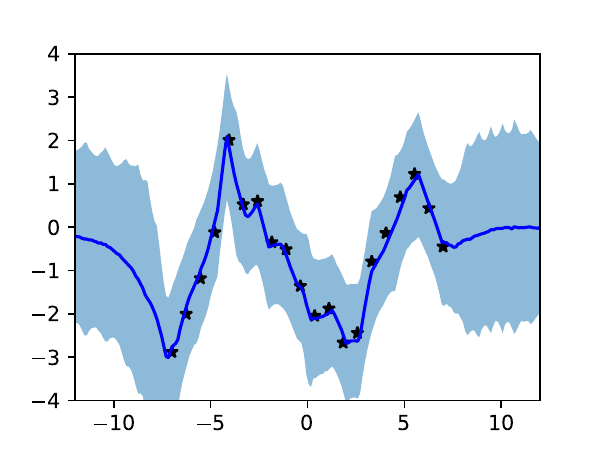}}
    \subfloat[\small{DNN.} \label{fig:nn1d}]{\includegraphics[width=.2\textwidth]{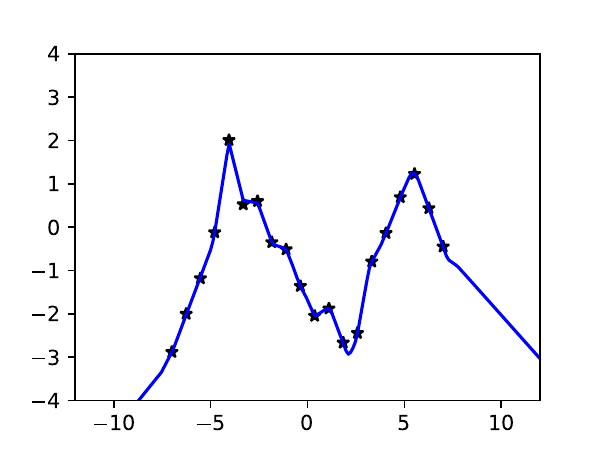}}
    \caption{\small{Results on toy dataset. (a)--(d) show the predictive posterior of the exact GP, DGP, exact DKL and proposed DAK model, respectively, on the noisy data generated by 1D GP with zero-mean and covariance function $k(x,x')=\exp( -(x-x')^2 )$. We set the number of MC samples $S=4$ for estimating the expected log-likelihood in ELBO during training.} The predictive mean and $\pm$2 standard deviations are plotted together with the observed data. (e) shows the NN fit with the same training data.}
    \label{fig:toyGP}
\end{figure*}

\section{RELATED WORK}
\label{sec:related work}
Motivated by integrating the power of deep networks with interpretable and theoretically grounded kernel methods, it is encouraging to see some contributions on such combinations across a range of contexts.

\paragraph{Variations of DKL.} 
Several recent studies have explored variations of DKL models. To handle large datasets and diverse tasks, \citet{wilson2016stochastic} extended the vanilla DKL \citep{wilson2016deep} to Stochastic Variational DKL (SV-DKL) by leveraging Stochastic Variational GP (SVGP) \citep{hensman2015scalable} and KISS-GP \citep{wilson2015kernel}. However, the kernel interpolation on a Cartesian grid does not scale in the high-dimensional space. \citet{xue2019deep} and \citet{xie2019deep} integrated deep kernel models with Random Fourier Features (RFF) \citep{rahimi2007random}, an efficient method for approximating GP kernel feature mappings. Some other work has focused on developing models with specialized kernel structures, such as recurrent kernels \citep{al2017learning} and compositional kernels \citep{sun2018differentiable}. More recently, 
\citet{ober2021promises} investigated overfitting in DKL models based on marginal likelihood maximization and proposed a fully Bayesian approach to mitigate this issue. \citet{matias2024amortized} introduced amortized variational DKL, which uses DNNs to learn variational distributions over inducing points in an amortized manner, thereby reducing overfitting by locally smoothing predictions. 

\paragraph{GPs and NNs.}
The connection between GPs and NNs was first established by \citet{neal1994bayesian}, who showed that the function defined by a single-layer NN with infinite width, random independent zero-mean weights, and biases is equivalent to a GP. This equivalence was later generalized to arbitrary NNs with infinite-width or infinite-depth layers in \citep{lee2017deep, cutajar2017random, matthews2018gaussian, yang2019wide, dutordoir2021deep, gao2023wide}. However, these studies focus primarily on fully connected NNs, which are not suitable for all practical applications. \citet{garriga2018deep} and \citet{novak2018bayesian} extended the equivalence to Convolutional Neural Networks (CNNs) \citep{lecun1989handwritten}, which are widely used in image recognition. Hybrid models combining GPs and NNs have also been investigated. \citet{bradshaw2017adversarial} proposed a hybrid GPDNN that feeds CNN features into a GP, while \citet{zhang2024gaussian} introduced GP Neural Additive Models (GP-NAM), a class of GAMs that use GPs approximated by RFF and a single-layer NN. However, GPDNN uses the standard GP, while we approximate GP via induced prior approximation. GP-NAM applies the additive model with RFF approximation but lacks Bayesian inference. \citet{harrison2024variational} introduced a sampling-free Bayesian last-layer architecture but did not establish a connection between this architecture and DKL. 

\section{EXPERIMENTS}
\label{sec:exp}

In this section, we evaluate the proposed DAK model on multiple real datasets for both regression tasks in \Cref{subsec:uci reg} and classification tasks in \Cref{subsec:image class}. We compare its performance with several baselines, including a neural network without GP integration (NN), DKL with SVGP \citep{titsias2009variational,hensman2015scalable} as the GP approximation (NN+SVGP), and SV-DKL \citep{wilson2016stochastic}. Furthermore, in \Cref{subsec:toy}, we present a toy example that demonstrates how the proposed model mitigates the issue of out-of-sample overfitting highlighted by \citet{ober2021promises}, in contrast to the NN model. 

\subsection{Toy Example: GP}
\label{subsec:toy}

\begin{table*}[ht]
\centering
\caption{\small{Comparison of regression performance on UCI datasets using 5-fold cross-validation with a batch size of 512 and a fully connected neural network architecture of $D \rightarrow 64 \rightarrow 32 \rightarrow D_{w}$, where the output features $D_{w}$ are 16, 64, and 256 respectively. The best results are highlighted in \textbf{bold}. Our models, DAK-MC (using MC approximation) and DAK-CF (using closed-form inference and ELBO), are highlighted with  \protect\colorbox{Gray}{gray background}.}}
\label{tab:uci metrics}
\vspace{-0.2cm}
\resizebox{\linewidth}{!}{%
\begin{tabular}{l|l|ccc|ccc|ccc}
\toprule[1pt]
\multirow{2}{*}{\makecell[tl]{\textbf{Dataset} \\ $(N, D)$ }}  & \multirow{2}{*}{\textbf{Model}}

& \multicolumn{3}{c|}{\textbf{NN out features = 16}} & \multicolumn{3}{c|}{\textbf{NN out features = 64}} & \multicolumn{3}{c}{\textbf{NN out features = 256}} \\
\cline{3-11}
&  & \textbf{RMSE} $\downarrow$ & \textbf{NLPD} $\downarrow$ & \textbf{Time (s)} $\downarrow$  & \textbf{RMSE} $\downarrow$ & \textbf{NLPD} $\downarrow$ & \textbf{Time (s)} $\downarrow$  & \textbf{RMSE} $\downarrow$ & \textbf{NLPD} $\downarrow$ & \textbf{Time (s)} $\downarrow$ \\
\hline
\multirow{5}{*}{ \makecell[tl]{ \textbf{Gas} \\ $(2565, 128)$} } & NN & 2.377 $\pm$ 2.399 & 4.749 $\pm$ 6.594 & \textbf{2.345 $\pm$ 0.003} & 2.107 $\pm$ 1.212 & 3.092 $\pm$ 2.229 & \textbf{2.362 $\pm$ 0.029} & 1.196 $\pm$ 0.712 & 1.730 $\pm$ 0.711 & \textbf{2.346 $\pm$ 0.004} \\
& NN+SVGP & 0.502 $\pm$ 0.171 & 1.121 $\pm$ 0.106 & 4.727 $\pm$ 0.009 & 0.625 $\pm$ 0.148 & 1.206 $\pm$ 0.119 & 4.724 $\pm$ 0.004 & 0.743 $\pm$ 0.183 & 1.322 $\pm$ 0.143 & 4.718 $\pm$ 0.009 \\
& SV-DKL & 0.589 $\pm$ 0.161 & 1.303 $\pm$ 0.227 & 28.189 $\pm$ 0.490 & 0.499 $\pm$ 0.183 & 1.235 $\pm$ 0.256 & 27.956 $\pm$ 0.078 & 0.534 $\pm$ 0.189 & 1.207 $\pm$ 0.209 & 28.400 $\pm$ 0.185 \\
& AV-DKL &  0.538 $\pm$ 0.129  &  1.250 $\pm$ 0.222  &  25.315 $\pm$ 0.254  & 0.604 $\pm$ 0.016  &  1.250 $\pm$ 0.010  &  27.070 $\pm$ 1.620 &  0.760 $\pm$ 0.367  &  1.276 $\pm$ 0.220  &  28.860 $\pm$ 1.903  \\
& \cellcolor{Gray}\raggedright DAK-MC & \cellcolor{Gray} \textbf{0.405 $\pm$ 0.061} & 
\cellcolor{Gray} \textbf{0.886 $\pm$ 0.048} & \cellcolor{Gray} 8.887 $\pm$ 0.007 & \cellcolor{Gray} 0.353 $\pm$ 0.046 & \cellcolor{Gray} \textbf{0.881 $\pm$ 0.053} & 
\cellcolor{Gray} 8.844 $\pm$ 0.005 & \cellcolor{Gray} 0.351 $\pm$ 0.019 & \cellcolor{Gray} \textbf{0.871 $\pm$ 0.027} & 
\cellcolor{Gray} 8.831 $\pm$ 0.017 \\
& \cellcolor{Gray}\raggedright DAK-CF & \cellcolor{Gray} 0.412 $\pm$ 0.134 & \cellcolor{Gray} 0.928 $\pm$ 0.100 & \cellcolor{Gray} 7.400 $\pm$ 0.004 & \cellcolor{Gray} \textbf{0.350 $\pm$ 0.020} & 
\cellcolor{Gray} 0.898 $\pm$ 0.040 & \cellcolor{Gray} 7.398 $\pm$ 0.009 & \cellcolor{Gray} \textbf{0.342 $\pm$ 0.033} & 
\cellcolor{Gray} 0.895 $\pm$ 0.046 & \cellcolor{Gray} 7.410 $\pm$ 0.009 \\
\hline
\multirow{5}{*}{ \makecell[tl]{ \textbf{Parkinsons} \\ $(5875, 20)$ } } & NN & 2.692 $\pm$ 1.302 & 3.503 $\pm$ 2.630 & \textbf{2.693 $\pm$ 0.027} & 2.288 $\pm$ 0.712 & 2.724 $\pm$ 1.158 & \textbf{2.589 $\pm$ 0.050} & 2.252 $\pm$ 0.757 & 2.720 $\pm$ 1.188 & \textbf{2.885 $\pm$ 0.026} \\
& NN+SVGP & 3.481 $\pm$ 1.906 & 4.606 $\pm$ 4.606 & 5.516 $\pm$ 0.117 & 3.238 $\pm$ 2.419 & 4.940 $\pm$ 5.494 & 5.467 $\pm$ 0.097 & 3.676 $\pm$ 3.287 & 6.791 $\pm$ 8.612 & 5.945 $\pm$ 0.090 \\
& SV-DKL & 2.608 $\pm$ 1.023 & 2.745 $\pm$ 1.097 & 35.804 $\pm$ 1.590 & 2.817 $\pm$ 1.670 & 3.193 $\pm$ 2.109 & 33.042 $\pm$ 0.306 & 2.896 $\pm$ 2.055 & 3.206 $\pm$ 1.906 & 32.882 $\pm$ 0.060 \\
& AV-DKL &  1.942 $\pm$ 0.758  &  \textbf{2.223 $\pm$ 0.655} &  30.337 $\pm$ 1.086  &  2.267 $\pm$ 0.584  &  \textbf{2.397 $\pm$ 0.628} & 31.006 $\pm$ 1.162 & 3.096 $\pm$ 0.472 & 3.105 $\pm$ 0.793 &  31.859 $\pm$ 1.197 \\
& \cellcolor{Gray}\raggedright DAK-MC & \cellcolor{Gray} 1.983 $\pm$ 1.154 & \cellcolor{Gray} 2.699 $\pm$ 1.819 & \cellcolor{Gray} 11.596 $\pm$ 0.260 & \cellcolor{Gray} 1.949 $\pm$ 0.912 & \cellcolor{Gray} 2.575 $\pm$ 1.121 & \cellcolor{Gray} 13.085 $\pm$ 0.055 & \cellcolor{Gray} 1.846 $\pm$ 0.974 & \cellcolor{Gray} 2.420 $\pm$ 1.212 & \cellcolor{Gray} 11.820 $\pm$ 0.296 \\
& \cellcolor{Gray}\raggedright DAK-CF & \cellcolor{Gray} \textbf{1.801 $\pm$ 1.013} & \cellcolor{Gray} 2.848 $\pm$ 1.810 & \cellcolor{Gray} 9.071 $\pm$ 0.083 & \cellcolor{Gray} \textbf{1.788 $\pm$ 0.997} & \cellcolor{Gray} 2.801 $\pm$ 1.853 & \cellcolor{Gray} 9.073 $\pm$ 0.048 & \cellcolor{Gray} \textbf{1.466 $\pm$ 1.093} & \cellcolor{Gray} \textbf{2.308 $\pm$ 1.892} & \cellcolor{Gray} 8.166 $\pm$ 0.071 \\
\hline
\multirow{5}{*}{ \makecell[tl]{\textbf{Wine} \\ $(1599, 11)$} } & NN & 0.728 $\pm$ 0.055 & 1.233 $\pm$ 0.057 & \textbf{2.350 $\pm$ 0.011} & 0.725 $\pm$ 0.057 & 1.231 $\pm$ 0.054 & \textbf{2.343 $\pm$ 0.007} & 0.712 $\pm$ 0.062 & 1.214 $\pm$ 0.062 & \textbf{2.367 $\pm$ 0.021} \\
& NN+SVGP & 0.739 $\pm$ 0.069 & 1.243 $\pm$ 0.067 & 4.713 $\pm$ 0.010 & 0.801 $\pm$ 0.130 & 1.325 $\pm$ 0.130 & 4.725 $\pm$ 0.008 & 0.893 $\pm$ 0.180 & 1.418 $\pm$ 0.147 & 4.718 $\pm$ 0.016 \\
& SV-DKL & 0.899 $\pm$ 0.110 & 1.443 $\pm$ 0.135 & 27.224 $\pm$ 0.170 & 0.947 $\pm$ 0.109 & 1.460 $\pm$ 0.112 & 27.136 $\pm$ 0.023 & 0.879 $\pm$ 0.130 & 1.434 $\pm$ 0.158 & 27.332 $\pm$ 0.101 \\
& AV-DKL & \textbf{0.699 $\pm$ 0.060} & 1.207 $\pm$ 0.068 & 24.419 $\pm$ 0.316 & \textbf{0.711 $\pm$ 0.055} & 1.221 $\pm$ 0.060 & 24.904 $\pm$ 0.424 & \textbf{0.711 $\pm$ 0.055} & 1.221 $\pm$ 0.060 & 24.904 $\pm$ 0.424 \\
& \cellcolor{Gray}\raggedright DAK-MC & \cellcolor{Gray} 0.756 $\pm$ 0.068 & \cellcolor{Gray} 1.164 $\pm$ 0.075 & \cellcolor{Gray} 8.820 $\pm$ 0.038 & \cellcolor{Gray} 0.751 $\pm$ 0.055 & \cellcolor{Gray} 1.163 $\pm$ 0.060 & \cellcolor{Gray} 8.821 $\pm$ 0.045 & \cellcolor{Gray} 0.727 $\pm$ 0.065 & \cellcolor{Gray} \textbf{1.140 $\pm$ 0.070} & \cellcolor{Gray} 8.765 $\pm$ 0.016 \\
& \cellcolor{Gray}\raggedright DAK-CF & \cellcolor{Gray}  0.736 $\pm$ 0.042 & \cellcolor{Gray} \textbf{1.162 $\pm$ 0.049} & \cellcolor{Gray} 7.376 $\pm$ 0.018 & \cellcolor{Gray} 0.728 $\pm$ 0.064 & \cellcolor{Gray} \textbf{1.153 $\pm$ 0.067} & 
\cellcolor{Gray} 7.352 $\pm$ 0.010 & 
\cellcolor{Gray} 0.720 $\pm$ 0.057 & \cellcolor{Gray} 1.147 $\pm$ 0.061 & 
\cellcolor{Gray} 7.415 $\pm$ 0.029 \\
\hline
\multirow{5}{*}{ \makecell[tl]{ \textbf{Kin40K} \\ $(40000, 8)$ }} & NN & 0.109 $\pm$ 0.016 & 0.752 $\pm$ 0.004 & \textbf{16.017 $\pm$ 0.114} & 0.100 $\pm$ 0.012 & 0.749 $\pm$ 0.003 & \textbf{16.912 $\pm$ 0.112} & 0.087 $\pm$ 0.016 & 0.746 $\pm$ 0.004 & \textbf{18.048 $\pm$ 0.025} \\
& NN+SVGP & 0.104 $\pm$ 0.013 & 0.751 $\pm$ 0.004 & 40.749 $\pm$ 0.508 & 0.142 $\pm$ 0.017 & 0.763 $\pm$ 0.006 & 41.239 $\pm$ 0.304 & 0.123 $\pm$ 0.016 & 0.757 $\pm$ 0.006 & 43.111 $\pm$ 1.482 \\
& SV-DKL & 0.096 $\pm$ 0.024 & 0.750 $\pm$ 0.010 & 230.284 $\pm$ 7.049 & 0.092 $\pm$ 0.018 & 0.748 $\pm$ 0.006 & 228.137 $\pm$ 7.406 & 0.097 $\pm$ 0.025 & 0.750 $\pm$ 0.009 & 226.606 $\pm$ 4.231 \\
& AV-DKL & 0.084 $\pm$ 0.011 & 0.746 $\pm$ 0.003 & 75.427 $\pm$ 0.965 &  0.081 $\pm$ 0.014 & 0.745 $\pm$ 0.003 & 87.654 $\pm$ 1.664 & 0.071 $\pm$ 0.015 & 0.743 $\pm$ 0.029 & 158.860 $\pm$ 3.621 \\
& \cellcolor{Gray}\raggedright DAK-MC & \cellcolor{Gray} 0.096 $\pm$ 0.042 & \cellcolor{Gray} 0.746 $\pm$ 0.010 & \cellcolor{Gray} 74.787 $\pm$ 0.383 & \cellcolor{Gray} 0.090 $\pm$ 0.028 & \cellcolor{Gray} 0.744 $\pm$ 0.006 & \cellcolor{Gray} 77.349 $\pm$ 5.409 & \cellcolor{Gray} 0.090 $\pm$ 0.031 & \cellcolor{Gray} 0.744 $\pm$ 0.007 & \cellcolor{Gray} 75.586 $\pm$ 0.313 \\
& \cellcolor{Gray}\raggedright DAK-CF & \cellcolor{Gray} \textbf{0.073 $\pm$ 0.015} & \cellcolor{Gray} \textbf{0.742 $\pm$ 0.005} & \cellcolor{Gray} 54.174 $\pm$ 0.322 & \cellcolor{Gray} \textbf{0.069 $\pm$ 0.018} & \cellcolor{Gray} \textbf{0.741 $\pm$ 0.005} & \cellcolor{Gray} 60.339 $\pm$ 0.161 & \cellcolor{Gray} \textbf{0.068 $\pm$ 0.015} & \cellcolor{Gray} \textbf{0.741 $\pm$ 0.004} & \cellcolor{Gray} 55.148 $\pm$ 0.165 \\
\hline
\multirow{5}{*}{ \makecell[tl]{ \textbf{Protein} \\ $(45730, 9)$} } & NN & 4.678 $\pm$ 7.804 & 25.091 $\pm$ 52.135 & \textbf{24.910 $\pm$ 0.221} & 0.906 $\pm$ 0.304 & 1.421 $\pm$ 0.232 & \textbf{19.433 $\pm$ 0.093} & 1.403 $\pm$ 1.408 & 2.358 $\pm$ 2.349 & \textbf{28.447 $\pm$ 0.133} \\
& NN+SVGP & 0.773 $\pm$ 0.003 & 1.278 $\pm$ 0.002 & 47.081 $\pm$ 0.643 & 0.773 $\pm$ 0.003 & 1.278 $\pm$ 0.002 & 38.986 $\pm$ 0.485 & 0.773 $\pm$ 0.003 & 1.278 $\pm$ 0.002 & 52.924 $\pm$ 1.144 \\
& SV-DKL & 0.769 $\pm$ 0.003 & 1.275 $\pm$ 0.001 & 262.274 $\pm$ 8.216 & 0.770 $\pm$ 0.003 & 1.276 $\pm$ 0.002 & 265.182 $\pm$ 3.680 & 0.769 $\pm$ 0.003 & 1.275 $\pm$ 0.001 & 264.678 $\pm$ 4.450 \\
& AV-DKL & 0.773 $\pm$ 0.003 & 1.278 $\pm$ 0.002 & 87.939 $\pm$ 1.303 &  0.773 $\pm$ 0.003 & 1.278 $\pm$ 0.002 & 92.591 $\pm$ 1.934 & 0.773 $\pm$ 0.003 & 1.278 $\pm$ 0.002 & 164.997 $\pm$ 4.108 \\
& \cellcolor{Gray}\raggedright DAK-MC & \cellcolor{Gray} \textbf{0.622 $\pm$ 0.047} & \cellcolor{Gray} \textbf{1.017 $\pm$ 0.043} & \cellcolor{Gray} 85.729 $\pm$ 1.035 & \cellcolor{Gray} \textbf{0.612 $\pm$ 0.038} & \cellcolor{Gray} \textbf{1.008 $\pm$ 0.035} & \cellcolor{Gray} 86.594 $\pm$ 0.827 & \cellcolor{Gray} \textbf{0.608 $\pm$ 0.042} & \cellcolor{Gray} \textbf{1.004 $\pm$ 0.039} & \cellcolor{Gray} 86.245 $\pm$ 0.088 \\
& \cellcolor{Gray}\raggedright DAK-CF & \cellcolor{Gray} 0.631 $\pm$ 0.021 & \cellcolor{Gray} 1.024 $\pm$ 0.019 & \cellcolor{Gray} 68.762 $\pm$ 0.379 & \cellcolor{Gray} 0.625 $\pm$ 0.027 & \cellcolor{Gray} 1.018 $\pm$ 0.026 & \cellcolor{Gray} 69.766 $\pm$ 0.190 & \cellcolor{Gray} 0.618 $\pm$ 0.029 & \cellcolor{Gray} 1.012 $\pm$ 0.026 & \cellcolor{Gray} 68.783 $\pm$ 0.222 \\
\hline
\multirow{5}{*}{ \makecell[tl]{ \textbf{KEGG} \\ $(48827, 20)$} } & NN & 0.132 $\pm$ 0.021 & 0.759 $\pm$ 0.006 & \textbf{21.856 $\pm$ 0.168} & 0.124 $\pm$ 0.009 & 0.758 $\pm$ 0.004 & \textbf{20.989 $\pm$ 0.132} & 0.127 $\pm$ 0.008 & 0.766 $\pm$ 0.019 & \textbf{23.143 $\pm$ 0.710} \\
& NN+SVGP & 0.128 $\pm$ 0.005 & 0.758 $\pm$ 0.001 & 43.788 $\pm$ 1.377 & 0.129 $\pm$ 0.013 & 0.759 $\pm$ 0.002 & 40.996 $\pm$ 0.280 & 0.127 $\pm$ 0.009 & 0.758 $\pm$ 0.002 & 46.882 $\pm$ 1.290 \\
& SV-DKL & 0.134 $\pm$ 0.011 & 0.766 $\pm$ 0.013 & 269.598 $\pm$ 5.733 & 0.139 $\pm$ 0.031 & 0.769 $\pm$ 0.025 & 271.083 $\pm$ 7.502 & 0.152 $\pm$ 0.024 & 0.780 $\pm$ 0.025 & 275.013 $\pm$ 2.443 \\
& AV-DKL & 0.127 $\pm$ 0.008 & 0.758 $\pm$ 0.002 & 92.000 $\pm$ 0.227 & 0.130 $\pm$ 0.009 & 0.759 $\pm$ 0.002 & 129.314 $\pm$ 1.607 & 0.130 $\pm$ 0.008 & 0.759 $\pm$ 0.002 & 254.537 $\pm$ 4.020 \\
& \cellcolor{Gray}\raggedright DAK-MC & \cellcolor{Gray} 0.126 $\pm$ 0.010 & \cellcolor{Gray} \textbf{0.748 $\pm$ 0.003} & \cellcolor{Gray} 90.473 $\pm$ 0.144 & \cellcolor{Gray} 0.124 $\pm$ 0.010 & \cellcolor{Gray} \textbf{0.748 $\pm$ 0.003} & \cellcolor{Gray} 92.507 $\pm$ 0.964 & \cellcolor{Gray} 0.123 $\pm$ 0.009 & \cellcolor{Gray} \textbf{0.748 $\pm$ 0.003} & \cellcolor{Gray} 113.738 $\pm$ 1.251 \\
& \cellcolor{Gray}\raggedright DAK-CF & \cellcolor{Gray} \textbf{0.124 $\pm$ 0.007} & \cellcolor{Gray} \textbf{0.748 $\pm$ 0.003} & \cellcolor{Gray} 65.070 $\pm$ 0.140 & \cellcolor{Gray} \textbf{0.122 $\pm$ 0.007} & \cellcolor{Gray} \textbf{0.748 $\pm$ 0.003} & \cellcolor{Gray} 79.732 $\pm$ 2.189 & \cellcolor{Gray} \textbf{0.121 $\pm$ 0.005} & \cellcolor{Gray} \textbf{0.748 $\pm$ 0.003} & \cellcolor{Gray} 78.702 $\pm$ 0.631 \\
\bottomrule[1pt]
\end{tabular}%
}
\end{table*}

We first evaluate a toy example of 1-Dimensional (1D) GP regression on synthetic data with zero mean and SE kernel $k(x,x')=\exp\left( -(x-x')^2 \right)$. The training set consists of $20$ noisy data points in the range of $[-7,7]$, while the test set consists of $100$ data points in $[-12,12]$. We consider a two-layer MLP with layer width $[64,32]$ as the feature extractor, letting $P=2$ be the number of \emph{base} GPs. To maintain a fair comparison, we use the same training recipe: full-batch training, a learning rate 0.01, and a number of optimization steps 1000. The only differences are the choice of the last layer and the loss function. We set the number of MC samples $S=4$ during training.

\Cref{fig:gp1d} -- \ref{fig:dak1d} shows the predictive posterior of the exact GP, two-layer DGP, exact DKL, and proposed DAK. We observe that DAK has good both in-sample and out-of-sample predictions, which is close to exact GP posterior, while DGP is unbiased but over-confident outside the training data. Exact DKL is biased and overconfident in out-of-sample predictions, which were also observed and investigated by \citet{ober2021promises}. We also compare DAK with the DNN that shares the same feature extractor and model depth. It is evident that the DNN fit in \Cref{fig:nn1d} suffers from ``overfitting'': the fit shows significant extrapolation beyond the training data.

\subsection{UCI Regression}
\label{subsec:uci reg}

We benchmark the regression task on six datasets from the UCI repository \citep{dua2017uci}: three smaller datasets with fewer than 10K samples (Wine, Gas, Parkinsons) and three larger datasets with 40K to 50K samples (Kin40K, Protein, KEGG). All models use the same neural network architecture: a fully connected network with two hidden layers of 64 and 32 neurons, respectively. Training is performed using the Adam optimizer over 100 epochs with a learning rate of $0.001$, weight decay of $0.0005$, and batch size of 512. The noise variance is set to $\sigma_{f}^2 = 0.01$.

For NN+SVGP, SV-DKL, and AV-DKL, we use SE kernels and 64 inducing points initialized with uniformly distributed random variables of a fixed seed. In the proposed DAK models, the induced grid $\Uv$ from \cref{eq:GPlayer} consists of $M = 7$ equally spaced points over the interval $(0,1)$ for each base GP, i.e., $\Uv = \{1/8, 2/8, \ldots, 7/8\}$. The NN outputs of both SV-DKL and our DAK model, which utilize an additive GP structure, are embedded and normalized into 16 base GPs over the interval $[0,1]$. That is, the number of projections $P$ in \cref{eq:deepadditive} is set to 16. For the DAK model, we evaluate two methods: DAK-CF, using the closed-form inference from \Cref{sec:uq of inference} and the closed-form ELBO from \Cref{sec:elbo}, and DAK-MC, using MC sampling to approximate the mean and variance during inference and approximate the expected log likelihood term in ELBO during training. The number of MC samples is set to 20 for inference and 8 for training, the same for NN+SVGP, SV-DKL, and AV-DKL. Further details can be found in \Cref{subsec:regression supp}.

We evaluate the performance of each model using training time (in seconds), Root Mean Squared Error (RMSE), and Negative Log Predictive Density (NLPD) with varying neural network output feature sizes of 16, 64, and 256. The results are averaged over 5-fold cross-validation for each dataset. As shown in \Cref{tab:uci metrics}, except for the smallest dataset, Wine, our models, DAK-MC and DAK-CF (highlighted in gray), consistently achieve the best RMSE and NLPD performance compared to other models. 

\begin{table*}[ht]
\caption{\small{ELBO, Accuracy, NLL, ECE for DAK, SV-DKL, NN+SVGP, NN on image classification tasks averaged over 3 runs. MNIST uses a simple CNN with 16 output features; CIFAR-10 uses ResNet-18 with 64 output features; CIFAR-100 uses ResNet-34 with 512 output features. The best results are highlighted in \textbf{bold}.}}
\centering
\vspace{-0.1cm}
\resizebox{\linewidth}{!}{%
\begin{tabular}{rcccclccc}
\toprule[1pt]
\multicolumn{1}{l}{} & \multicolumn{4}{c}{Batch size: 128}  &  & \multicolumn{3}{c}{Batch size: 1024} \\ \cline{2-5} \cline{7-9} \vspace{-8pt} \\
\multicolumn{1}{l}{} & NN   & NN+SVGP & SV-DKL & \cellcolor{Gray} DAK-MC &   & NN+SVGP  & SV-DKL & \cellcolor{Gray} DAK-MC \\ 
\midrule[1pt]
MNIST - ELBO $\uparrow$        & \rule{0.5cm}{0.4pt}  &  -0.121 $\pm$ 0.000    & -0.054 $\pm$ 0.000  & \cellcolor{Gray} \textbf{-0.030 $\pm$ 0.001}   &     & -1.997 $\pm$ 0.001        &  -0.305 $\pm$ 0.001      &  \cellcolor{Gray} \textbf{-0.048 $\pm$ 0.001}  \\
Top-1 Acc. (\%) $\uparrow$       & 99.14 $\pm$ 0.00     & 98.56 $\pm$ 0.02       & 99.16 $\pm$ 0.00  & \cellcolor{Gray} \textbf{99.26 $\pm$ 0.01}      &      & 14.79 $\pm$ 0.08         & 96.23 $\pm$ 0.02       &  \cellcolor{Gray} \textbf{99.1 $\pm$ 0.00}    \\
NLL $\downarrow$             & 0.026 $\pm$ 0.000    & 0.064 $\pm$ 0.001     & 0.030 $\pm$ 0.000      & \cellcolor{Gray} \textbf{0.024 $\pm$ 0.000}   &    & 1.985 $\pm$ 0.003        &  0.288 $\pm$ 0.001      &  \cellcolor{Gray} \textbf{0.028 $\pm$ 0.004}      \\
ECE $\downarrow$          & 0.005 $\pm$ 0.000     & 0.012 $\pm$ 0.000    & 0.005 $\pm$ 0.000       & \cellcolor{Gray} \textbf{0.004 $\pm$ 0.000}           &      & 0.062 $\pm$ 0.002         & 0.020 $\pm$ 0.000       & \cellcolor{Gray} \textbf{0.006 $\pm$ 0.000} \\
\midrule[1pt]
CIFAR-10 - ELBO $\uparrow$      & \rule{0.5cm}{0.4pt}     & -1.038 $\pm$ 0.004        & -0.017 $\pm$  0.001      &  \cellcolor{Gray} \textbf{-0.002 $\pm$ 0.000}     &     & -1.039 $\pm$ 0.004         & -0.034 $\pm$ 0.021       & \cellcolor{Gray} \textbf{-0.002 $\pm$ 0.000} \\
Top-1 Acc. (\%) $\uparrow$    & 94.72 $\pm$ 0.13   & 77.35 $\pm$ 0.13  & 93.44 $\pm$ 0.28    &  \cellcolor{Gray} \textbf{94.81 $\pm$ 0.13}   &     &  16.90 $\pm$ 0.29        & 90.22 $\pm$ 1.42       & \cellcolor{Gray} \textbf{93.02 $\pm$ 0.18}        \\
NLL $\downarrow$     & \textbf{0.252 $\pm$ 0.025}      & 1.790 $\pm$ 0.001    & 0.312 $\pm$ 0.033       &  \cellcolor{Gray} 0.256 $\pm$ 0.014     &      & 2.270 $\pm$ 0.000         & 0.485 $\pm$ 0.061       & \cellcolor{Gray} \textbf{0.345 $\pm$ 0.001}    \\
ECE $\downarrow$      & 0.040 $\pm$ 0.002     & 0.061 $\pm$ 0.003    & 0.046 $\pm$ 0.003       &  \cellcolor{Gray} \textbf{0.039 $\pm$ 0.002}          &     & \textbf{0.027 $\pm$ 0.002}       & 0.060 $\pm$ 0.004       & \cellcolor{Gray} 0.052 $\pm$ 0.001           \\
\midrule[1pt]
CIFAR-100 - ELBO $\uparrow$      & \rule{0.5cm}{0.4pt}     & -4.605 $\pm$ 0.000        & -0.059 $\pm$ 0.000       &  \cellcolor{Gray} \textbf{-0.003 $\pm$ 0.000}      &     & -4.605 $\pm$ 0.000         & -0.103 $\pm$ 0.009       & \cellcolor{Gray} \textbf{-0.005 $\pm$ 0.001}   \\
Top-1 Acc. (\%) $\uparrow$    & 75.88 $\pm$ 0.54   & 1.04 $\pm$ 0.01  & 74.52 $\pm$ 0.13       & \cellcolor{Gray}  \textbf{76.75 $\pm$ 0.18}     &     &  1.10 $\pm$ 0.09        & 66.54 $\pm$ 0.74       & \cellcolor{Gray} \textbf{70.38 $\pm$ 1.25}        \\
NLL $\downarrow$     & 1.018 $\pm$ 0.021      & 4.605 $\pm$ 0.000    & 1.041 $\pm$ 0.007       & \cellcolor{Gray}  \textbf{1.001 $\pm$ 0.027}     &      & 4.605 $\pm$ 0.000    & 1.738 $\pm$  0.058      & \cellcolor{Gray} \textbf{1.203 $\pm$ 0.040}        \\
ECE $\downarrow$      & 0.086 $\pm$ 0.002     & \textbf{0.003 $\pm$ 0.001}    & 0.049 $\pm$ 0.002       & \cellcolor{Gray}  0.041 $\pm$ 0.004        &     & \textbf{0.003 $\pm$ 0.001}         & 0.148 $\pm$ 0.007       &\cellcolor{Gray}  0.056 $\pm$ 0.006           \\
\bottomrule[1pt]
\end{tabular}
}
\label{tab:img metrics}
\end{table*}

Although NN+SVGP takes less time than DAK models, its performance often degrades, with higher RMSE and NLPD as the size of the NN output features increases. In contrast, the DAK models show improved performance with larger NN output features. This is because SVGP has a fixed number of 64 inducing points, which limits its ability to approximate GPs effectively in high-dimensional spaces, making it less suitable for complex tasks requiring high-dimensional NNs, such as multitask regression or meta-learning.

Additionally, SV-DKL, which also uses an additive GP layer and KISS-GP to accelerate GP computations, takes significantly more time than our DAK models. This is because SV-DKL treats dependent inducing variables as parameters in VI, which requires more training time, while our DAK models treat the GP layers as sparse BNNs with independent Gaussian weights and biases under the mean-field assumption, leading to more efficient training. 

While AV-DKL occasionally achieves higher accuracy, it also requires substantially more training time than our DAK methods. By using inducing locations dependent on NN outputs to mitigate overcorrelation in DKL, AV-DKL significantly increases the ELBO’s computational complexity. Similar to NN+SVGP, its performance degrades when the size of the NN output features grows, because GP approximation in higher-dimensional spaces would naturally require a corresponding increase in the number of inducing points.

\subsection{Image Classification}
\label{subsec:image class}

\begin{table*}[ht]
\caption{\small{The number of total trainable parameters and average training time per epoch across different tasks for each model. NN has smaller set of parameters and less training time as is expected. DAK is more scalable than SV-DKL in terms of total trainable parameters and training time per epoch.}}
\centering
\vspace{-0.1cm}
\resizebox{\linewidth}{!}{%
\begin{tabular}{lllllccccccccc}
\toprule[1pt]
\multirow{2}{*}{Datasets} & \multirow{2}{*}{$N$} & \multirow{2}{*}{$D$} & \multirow{2}{*}{$C$} & \multirow{2}{*}{$D_w$} & \multicolumn{4}{c}{\# parameters} &  & \multicolumn{4}{c}{Epoch time (sec.)} \\ \cline{6-9} \cline{11-14} \vspace{-8pt} \\
                          &                    &                    &                    &                    & NN  & NN+SVGP  & SV-DKL  & \cellcolor{Gray} DAK-MC &  & NN & NN+SVGP & SV-DKL & \cellcolor{Gray} DAK-MC \\
\midrule[1pt]
MNIST                     & 60K                & 28$\times$28                 & 10                 & 16                 & 1.19M    &  $+$2.72M        &  $+$0.08M       & \cellcolor{Gray}  \textbf{$+$0.03M}      &  & 6.18   &  $+$16.57       &  \textbf{$+$4.35}      & \cellcolor{Gray}  $+$4.57      \\
CIFAR-10                  & 50K                & 3$\times$32$\times$32                 & 10                 & 64                 & 11.17M    &  $+$29.58M        & $+$0.30M        & \cellcolor{Gray}  \textbf{$+$0.11M}      &  & 34.41   &  $+$22.92       &  $+$7.88      & \cellcolor{Gray}  \textbf{$+$5.02}      \\
CIFAR-100                 & 50K                & 3$\times$32$\times$32                 & 100                & 512                & 21.32M    &  $+$52.53M        &  $+$2.19M       & \cellcolor{Gray}  \textbf{$+$1.68M}     &  & 42.82   &  $+$101.77       & $+$16.61       & \cellcolor{Gray}  \textbf{$+$6.24}     \\
\bottomrule[1pt]
\end{tabular}
}
\label{tab:runtime}
\end{table*}

We next benchmark the classification task on high-dimensional and highly structured image data, including MNIST \citep{lecun1998mnist}, CIFAR-10, and CIFAR-100 \citep{krizhevsky2009learning}. All models share the same neural network head as feature extractors, to which we add either a linear output layer in NN model or corresponding GP output layers in NN+SVGP, SV-DKL, and DAK. In classification, last layer outputs are followed by a Softmax layer to normalize the output to a probability distribution, and we perform MC sampling to approximate the Softmax likelihood term in ELBO. 

We use the same setting of training for all models (refer to \Cref{tab:optimizer classification} in \Cref{subsec:classification supp} for details). For MNIST, we consider a simple CNN as the feature extractor, 20 epochs of training using Adadelta optimizer with initial learning rate 1.0, weight decay 0.0001, and annealing learning rate scheduler at each step with a factor of 0.7. For CIFAR-10, we perform full-training on a ResNet-18 \citep{he2016deep} with GP layers for 200 epochs, while for CIFAR-100, we use a pre-trained ResNet-34 as the feature extractor and fine-tune GP layers for 200 epochs since NN+SVGP and SV-DKL struggled to fit. In both CIFAR10/100, we use SGD optimizer with an initial learning rate of 0.1, weight decay of 0.0001, momentum 0.9, and cosine annealing learning rate scheduler. 

For NN+SVGP, we use SE kernel and $512$ inducing points, while for SV-DKL, we use $64$ inducing points initialized with uniformly distributed random variable within the interval $[-1,1]^{D_w}$. For the proposed DAK-MC, we use $M=63$ equally spaced points over the interval $[-1,1]$ for the induced grid $\Uv$ in each base GP. More details can be found in \Cref{subsec:classification supp}.

We evaluate all models in terms of Top-1 accuracy, NLL, ELBO, and expected calibration error (ECE) over three runs. As shown in \Cref{tab:img metrics}, our DAK-MC (highlighted in gray) achieves the best accuracy, ELBO, and NLL performance compared to other DKL models. Although some ECEs of NN+SVGP and SV-DKL are lower than those of DAK-MC, their accuracy degrades more with increasing dimensionality. The failure of SVGP in CIFAR-10/100 demonstrated the necessity of additive structure in high-dimensional multitask DKL. Additionally, we observe that SVDKL struggles more to fit in CIFAR-100, indicating the importance of pre-training in SVDKL, while our proposed DAK does not hurt by the curse of dimensionality because of the computational advantages of the last-layer BNN feature. We also repeated the experiments with a larger batch size of 1024. Our proposed DAK model is more robust than other DKL methods when the batch size and the number of features increases. Further experimental results are provided in \Cref{sec:additional exp results}. In \Cref{tab:runtime}, we measure the total number of trainable parameters and the average training time. NN has the smallest set of parameters and the least training time as expected. Among DKL methods, DAK is more efficient than SV-DKL when large-scale neural networks and high-dimensional tasks are applied.

\section{CONCLUSION}
\label{sec:conc}
In this work, we introduced the DAK model, which reinterprets DKL as a hybrid NN, representing the last-layer GP as
a sparse BNN layer to address the scalability and training inefficiencies of DKL. 
The DAK model overcomes limitations of GPs by embedding high-dimenional features from NNs into additive GP layers and leveraging the sparse Cholesky factor of the Laplace kernel on the induced grids, significantly enhancing training and inference efficiency. The proposed model also provides closed-form solutions for both the predictive distribution and ELBO in regression tasks, eliminating the need for costly MC sampling. Empirical results show that DAK outperforms state-of-the-art DKL models in both regression and classification tasks. This work also opens new possibilities for improving DKL by establishing the connection between BNNs and GPs. 

Possible directions for future work include considering more general GP layers other than the Laplace kernels and the additive structure, and exploring other variational families for training the BNNs.

\ifthenelse{\equal{\showcontent}{1}}{
\subsubsection*{Acknowledgements}
The authors gratefully acknowledge the support provided by NSF grant DMS-2312173, the computing infrastructure provided by the Department of Electrical \& Computer Engineering at Texas A\&M University, and the reviewers' constructive feedback.}

\bibliographystyle{apalike}
\bibliography{references}

\section*{Checklist}
 \begin{enumerate}

 \item For all models and algorithms presented, check if you include:
 \begin{enumerate}
   \item A clear description of the mathematical setting, assumptions, algorithm, and/or model. [\textbf{Yes}/No/Not Applicable]
   \item An analysis of the properties and complexity (time, space, sample size) of any algorithm. [\textbf{Yes}/No/Not Applicable]
   \item (Optional) Anonymized source code, with specification of all dependencies, including external libraries. [\textbf{Yes}/No/Not Applicable]
 \end{enumerate}

 \item For any theoretical claim, check if you include:
 \begin{enumerate}
   \item Statements of the full set of assumptions of all theoretical results. [\textbf{Yes}/No/Not Applicable]
   \item Complete proofs of all theoretical results. [\textbf{Yes}/No/Not Applicable]
   \item Clear explanations of any assumptions. [\textbf{Yes}/No/Not Applicable]     
 \end{enumerate}

 \item For all figures and tables that present empirical results, check if you include:
 \begin{enumerate}
   \item The code, data, and instructions needed to reproduce the main experimental results (either in the supplemental material or as a URL). [\textbf{Yes}/No/Not Applicable]

   \item All the training details (e.g., data splits, hyperparameters, how they were chosen). [\textbf{Yes}/No/Not Applicable]
         \item A clear definition of the specific measure or statistics and error bars (e.g., with respect to the random seed after running experiments multiple times). [\textbf{Yes}/No/Not Applicable]
         \item A description of the computing infrastructure used. (e.g., type of GPUs, internal cluster, or cloud provider). [\textbf{Yes}/No/Not Applicable]
 \end{enumerate}

 \item If you are using existing assets (e.g., code, data, models) or curating/releasing new assets, check if you include:
 \begin{enumerate}
   \item Citations of the creator If your work uses existing assets. [\textbf{Yes}/No/Not Applicable]
   \item The license information of the assets, if applicable. [Yes/No/\textbf{Not Applicable}]
   \item New assets either in the supplemental material or as a URL, if applicable. [Yes/No/\textbf{Not Applicable}]
   \item Information about consent from data providers/curators. [Yes/No/\textbf{Not Applicable}]
   \item Discussion of sensible content if applicable, e.g., personally identifiable information or offensive content. [Yes/No/\textbf{Not Applicable}]
 \end{enumerate}

 \item If you used crowdsourcing or conducted research with human subjects, check if you include:
 \begin{enumerate}
   \item The full text of instructions given to participants and screenshots. [Yes/No/\textbf{Not Applicable}]
   \item Descriptions of potential participant risks, with links to Institutional Review Board (IRB) approvals if applicable. [Yes/No/\textbf{Not Applicable}]
   \item The estimated hourly wage paid to participants and the total amount spent on participant compensation. [Yes/No/\textbf{Not Applicable}]
 \end{enumerate}

 \end{enumerate}


%
%





%

%

\onecolumn
\appendix
\aistatstitle{From Deep Additive Kernel Learning to Last-Layer \\ Bayesian Neural Networks via Induced Prior Approximation: \\
Supplementary Materials}

\section{SPARSE CHOLESKY DECOMPOSITION}
\label{sec:sparse chol decompose}
In this section, we present the algorithm for constructing the induced grids $\mathbf{U}$ as defined in \cref{eq:GPlayer} by using sorted dyadic points, and obtaining the sparse Choleksy decomposition of the Laplace kernel in one dimension, as proposed in \citep{ding2024sparse}.

A set of one-dimensional level-$L$ dyadic points $\Xv_L$ in increasing order over the interval $[0,1]$ is defined as:
\begin{align}
    \Xv_{L}:= \left\{ \frac{1}{2^{L}}, \frac{2}{2^{L}}, \frac{3}{2^{L}}, \ldots, \frac{2^{L}-1}{2^{L}} \right\}.
\end{align}
However, this increasing order does not yield a sparse representation of the Markov kernel $k(\cdot,\cdot)$ on the points $\Xv_L$, i.e., Cholesky decomposition of the covariance matrix $k(\Xv_L, \Xv_L)$ is not sparse. To achieve a sparse hierarchical expansion, we first sort the dyadic points $\Xv_L$ according to their levels.

\paragraph{Sorted Dyadic Points}
For level-$\ell$ dyadic points $\Xv_{\ell}$ where $ \ell=1,\ldots,L$, we first define the set $\rho(\ell)$ consisting of odd numbers as follows:
\begin{align}
    \rho(\ell) = \left\{ 1,3,5,\ldots,2^{\ell}-1 \right\}.
\end{align}
Next, we define the sorted incremental set $\Dv_{\ell}$ (with $\Xv_{0}:= \varnothing$) as:
\begin{align}
    \Dv_{\ell} = 
    \left\{ \frac{i}{2^{\ell}}: i\in \rho(\ell) \right\} = \Xv_{\ell} - \Xv_{\ell-1}, \quad  \ell=1,\ldots L.
\end{align}
Thus, the level-$L$ dyadic points $\Xv_L$ can be decomposed into disjoint incremental sets $\{ \Dv_{\ell} \}_{\ell=1}^{L}$:
\begin{align}
    \Xv_{L} = \cup_{\ell=1}^{L} \Dv_{\ell}, \quad \Dv_{i} \cap \Dv_{j} = \varnothing \text{ for $i\neq j$}.
\end{align}
Therefore, we can define the sorted level-$L$ dyadic points using these incremental sets as:
\begin{align}\label{eq:sorted dyadic}
    \Xv_{L}^{\text{sort}}:= \left\{ \Dv_1,\Dv_2, \ldots, \Dv_{L} \right\} 
    = \left\{ \frac{i \in \rho(\ell) }{2^{\ell}}, \ell=1,\ldots,L \right\}.
\end{align}
For example, the sorted level-3 dyadic points are given by:
\begin{align}
    \Xv_{3}^{\text{sort}} 
    = \bigg\{ 
    \begingroup
        \color{blue}
        \underbracket{
            \color{black}
            \frac{1}{2^1}
        }_{\color{blue}
            \Dv_1
        }
    \endgroup
    , 
    \begingroup
        \color{blue}
        \underbracket{
            \color{black}
            \frac{1}{2^2}, \frac{3}{2^2}
        }_{\color{blue}
            \Dv_2
        }
    \endgroup
    ,
    \begingroup
        \color{blue}
        \underbracket{
            \color{black}
            \frac{1}{2^3}, \frac{3}{2^3}, \frac{5}{2^3}, \frac{7}{2^3}
        }_{\color{blue}
            \Dv_3
        }
    \endgroup
     \bigg\}.
\end{align}

\paragraph{Algorithm}
We now present the algorithm for computing the inverse of the upper triangular Cholesky factor $[ \Lv_{\Xv_{L}^{\text{sort}}}^{\top} ]^{-1}$ of the covariance matrix $k(\Xv_{L}^{\text{sort}}, \Xv_{L}^{\text{sort}})$ in \Cref{alg:cholesky}, where $\Lv_{\Xv_{L}^{\text{sort}}} \Lv_{\Xv_{L}^{\text{sort}}}^{\top} = k(\Xv_{L}^{\text{sort}}, \Xv_{L}^{\text{sort}})$.. The corresponding proof can be found in \citep{ding2024sparse}. The output of \Cref{alg:cholesky} is a sparse matrix with $\Oc(3 \cdot (2^{L}-1))$ nonzero entries. Since each iteration of the for-loop only requires solving a $3 \times 3$ linear system, which costs $\Oc(3^3)$ time, the total computational complexity of \Cref{alg:cholesky} is $\Oc(2^L-1)$. This implies that the complexity of computing $\left[ \Lv_{\Uv}^{\top} \right]^{-1}$ in \cref{eq:GPlayer} is $\Oc(M)$ when $\Uv$, the induced grid of size $M$, consists of sorted dyadic points as defined in \cref{eq:sorted dyadic}.

\begin{algorithm}[hbt!]
\caption{Computation of the inverse Cholesky factor for the Markov kernel $k(\cdot, \cdot)$ on sorted one-dimensional level-$L$ dyadic points $\Xv_L^{\text{sort}}$.}
\label{alg:cholesky}
\setstretch{0.99} 
\begin{algorithmic}[1]
    \STATE {\bfseries Input:} Markov kernel $k(\cdot,\cdot)$, sorted level-$L$ dyadic points $\Xv_{L}^{\text{sort}}$
    \STATE {\bfseries Output:} inverse of the upper triangular Cholesky factor $\Rv:= [ \Lv_{\Xv_{L}^{\text{sort}}}^{\top} ]^{-1}$, s.t. $\Lv_{\Xv_{L}^{\text{sort}}} \Lv_{\Xv_{L}^{\text{sort}}}^{\top} = k(\Xv_{L}^{\text{sort}}, \Xv_{L}^{\text{sort}})$
    \STATE Initialize $\Rv \leftarrow \text{zeros($2^L-1$,$2^L-1$)}$;
    \STATE Define $k(\pm \infty, \cdot) = k(\cdot, \pm \infty) = 0$;
    \FOR{$\ell=1$ {\bfseries to} $L$}
        \FOR{$i \in \rho(\ell)=\{1,3,\ldots,2^{\ell}-1\}$}
            \STATE $x_{\text{mid}} := \frac{i}{2^{\ell}}$;\quad
            $x_{\text{left}}:=\frac{i-1}{2^{\ell}}$ {\bfseries if} $i>1$ {\bfseries else} $-\infty$;\quad
            $x_{\text{right}}:=\frac{i+1}{2^{\ell}}$ {\bfseries if} $i<2^{\ell}-1$ {\bfseries else} $+\infty$;
            \STATE Get $i_{\text{mid}}$, $i_{\text{left}}$, $i_{\text{right}}$, the indices of the points $x_{\text{mid}}$, $x_{\text{left}}$, $x_{\text{right}}$ in the sorted set $\Xv_{L}^{\text{sort}}$ respectively;
            \STATE Get the coefficients $c_1$, $c_2$, $c_3$ by solving the following linear system:
            \begin{align}
                \begin{bmatrix}
                     & k(x_{\text{left}}, x_{\text{left}})
                     & k(x_{\text{left}}, x_{\text{mid}})
                     & k(x_{\text{left}}, x_{\text{right}}) \\
                     & k(x_{\text{mid}}, x_{\text{left}})
                     & k(x_{\text{mid}}, x_{\text{mid}})
                     & k(x_{\text{mid}}, x_{\text{right}}) \\
                     & k(x_{\text{right}}, x_{\text{left}})
                     & k(x_{\text{right}}, x_{\text{mid}})
                    &k(x_{\text{right}}, x_{\text{right}})
                \end{bmatrix}
                \begin{bmatrix}
                    c1\\
                    c2\\
                    c3
                \end{bmatrix}=
                \begin{bmatrix}
                    0\\
                    1\\
                    0
                \end{bmatrix}.
            \end{align}
            \STATE $[c_1,c_2,c_3] := [c_1,c_2,c_3] / \sqrt{c_2}$;
            \STATE {\bfseries if} $x_{\text{left}} \neq - \infty$, 
            {\bfseries then} $\Rv[i_{\text{left}} ,i_{\text{mid}}] = c_1$; \quad
            {\bfseries if} $x_{\text{right}} \neq + \infty$, 
            {\bfseries then} $\Rv[i_{\text{right}} ,i_{\text{mid}}] = c_3$;
            \STATE $\Rv[i_{\text{mid}} ,i_{\text{mid}}] = c_2$;
        \ENDFOR
    \ENDFOR
\end{algorithmic}
\end{algorithm}

\section{REPARAMETERIZATION OF KERNEL LENGTHSCALES}
\label{sec:theo}
Considering the additive Laplace kernel with fixed lengthscale $\tilde{\theta}$ for all base kernels, applying linear projections $\left\{ \wv_{p}^{\top}\xv \right\}_{p=1}^{P}$ on inputs $\xv\in \Rb^D$ will give:
\begin{align}
    &\sum_{p=1}^{P}\sigma^2_p k_p\left( \wv^{\top}_{p}\xv,\wv^{\top}_{p}\xv^{\prime} \right)\nonumber \\
    = & \sum_{p=1}^{P} \sigma^2_p\exp \left( -  \frac{\sum_{d=1}^{D} \left| w_{p,d}\left( x_{d}-x_{d}^{\prime} \right) \right|}{\tilde{\theta}} \right)\nonumber \\
    = & \sum_{p=1}^{P} \prod_{d=1}^{D} \sigma^2_p\exp \left( - \frac{\left| x_{d}-x_{d}^{\prime} \right|}{\tilde{\theta} / \left| w_{p,d}\right| } \right)\nonumber \\
    = & \sum_{p=1}^{P} \prod_{d=1}^{D} \sigma^2_p\exp \left( - \frac{\left| x_{d}-x_{d}^{\prime} \right|}{\theta_{p,d}} \right),
\end{align}
This still leads to an additive Laplace kernel but with adaptive lengthscale $\theta_{p,d}$ for base kernels. The resulting kernel also retains \emph{sparse} Cholesky decomposition by the properties of Markov kernels so that the complexity of inference is $\Oc(M)$.

\section{INFERENCE OF PREDICTIVE DISTRIBUTION}
\label{sec:uq of inference}
Given an input $\xv \in \Rb^D$, the prediction of the DAK model can be written in the following equation according to \cref{eq:DAK prediction}: 
\begin{align}
    \tilde{f}_{\xv}
    &= \sum_{p=1}^{P}
    \sigma_p \Big(
        \phi(h_{\psi}^{[p]}(\xv)) \zv_p
    \Big) + \mu \nonumber\\
    &= \sum_{p=1}^{P}
    \sigma_p \Big(
        \bm{\phi}_{p}^{\top} \zv_p
    \Big) + \mu,
\end{align}
where $\bm{\phi}_{p}^{\top}:=\phi(h_{\psi}^{[p]}(\xv)) \in \Rb^{1 \times M}$
. We assume the variational distribution over the independent Gaussian weights $\zv_p \sim \Nc(\bm{m}_{\zv_p}, \Sv_{\zv_p})$ and the bias $\mu \sim \Nc(m_{\mu}, \sigma_{\mu}^2)$. Then it's straighforward to deduce that
\begin{align}
    \bm{\phi}_{p}^{\top} \zv_p + \mu 
    &\sim
    \Nc\left(
    \bm{\phi}_{p}^{\top} \bm{m}_{\zv_p} + m_{\mu},\hspace{0.2em}
    \bm{\phi}_{p}^{\top} \Sv_{\zv_p} \bm{\phi}_{p} + \sigma_{\mu}^2
    \right), \\
    \sigma_p \left(
    \bm{\phi}_{p}^{\top} \zv_p 
    \right) + \mu
    & \sim
    \Nc\left(
    \sigma_p ( \bm{\phi}_{p}^{\top} \bm{m}_{\zv_p} )+ m_{\mu} ,\hspace{0.2em}
    \sigma_p^2( \bm{\phi}_{p}^{\top} \Sv_{\zv_p} \bm{\phi}_{p}) + \sigma_{\mu}^2
    \right), \\
    \tilde{f}_{\xv} = 
    \sum_{p=1}^{P}
    \sigma_p \left(
    \bm{\phi}_{p}^{\top} \zv_p
    \right) + \mu
    & \sim
    \Nc\left(
    \sum_{p=1}^{P}
    \sigma_p ( \bm{\phi}_{p}^{\top} \bm{m}_{\zv_p}) + m_{\mu} ,\hspace{0.2em}
    \sum_{p=1}^{P}
    \sigma_p^2( \bm{\phi}_{p}^{\top} \Sv_{\zv_p} \bm{\phi}_{p} ) + \sigma_{\mu}^2
    \right).
\end{align}
Therefore, we obtain the predictive distribution of the $\tilde{f}(\xv)$ at the point $\xv \in \Rb^D$ and its mean and variance are given by:
\begin{subequations}
\label{eq:dak inference closed form}
\begin{align}
    \Eb\left[ \tilde{f}_{\xv} \right]
        = \sum_{p=1}^{P}
        \sigma_p ( \bm{\phi}_{p}^{\top} \bm{m}_{\zv_p}) + m_{\mu},
\end{align}
\begin{align}
    \text{Var}\left[ \tilde{f}_{\xv} \right]
        =\sum_{p=1}^{P}
        \sigma_p^2( \bm{\phi}_{p}^{\top} \Sv_{\zv_p} \bm{\phi}_{p}) + \sigma_{\mu}^2.
\end{align}
\end{subequations}

\section{TRAINING OF VARIATIONAL INFERENCE}
\label{sec:training}
Given the dataset $\mathcal{D}=\{ \Xv, \yv \}$ where $\Xv:=\{ \xv_i \}_{i=1}^N$, $\yv=(y_1,\ldots,y_N)^{\top}$, $\xv_i \in \Rb^D$, $y_i\in\Rb$, the prediction $\tilde{f}_{\Xv}\in \Rb^N$ of DAK is given by all the parameters $\bm{\theta}=\left\{ \psi, \bm{\sigma} \right\}$, $\bm{\eta}=\left\{ \{ \mv_{\zv_{p}},\Sv_{\zv_{p}}\}_{p=1}^{P} , \{m_{\mu},\sigma_{\mu} \} \right\}$ according to \cref{eq:DAK prediction}:
\begin{align}
    \tilde{f}_{\Xv}:= \tilde{f}(\Xv; \bm{\theta}, \bm{\eta})
    = \sum_{p=1}^{P}
    \sigma_p \Big(
        \phi(h_{\psi}^{[p]}(\Xv)) \zv_p
    \Big) + \mu,
\end{align}
where $\zv_{p} \sim \mathcal{N} (\bm{m}_{\zv_p} ,\Sv_{\zv_p})$, $p=1,\ldots,P$, and $\mu \sim \mathcal{N} ( m_{\mu},\sigma^2_{\mu} )$ are variational variables $\Theta_{\text{var}}$ parameterized by $\bm{\eta}$. The variational distribution is denoted by $q_{\bm{\eta}}(\Theta_{\text{var}})= q(\mu)\prod_{p=1}^{P} q(\zv_{p}) = \Nc ( m_{\mu} ,\sigma_{\mu}^2 )\prod_{p=1}^{P} 
\Nc ( \bm{m}_{\zv_p} ,\Sv_{\zv_p} )$, and the variational prior is denoted by $p(\Theta_{\text{var}})$.

We consider the KL divergence between $q_{\bm{\eta}}(\Theta_{\text{var}})$ and the true posterior $p(\Theta_{\text{var}}\vert \yv, \Xv, \bm{\theta})$:
\begin{align}
& \qquad \text{KL} \left[ q_{\bm{\eta}}(\Theta_{\text{var}}) \| p(\Theta_{\text{var}} \vert \yv,\Xv, \bm{\theta} ) \right] \nonumber \\
= & \int q_{\bm{\eta}}(\Theta_{\text{var}} )\log \frac{q_{\bm{\eta}}(\Theta_{\text{var}} )}{p(\Theta_{\text{var}} \vert \yv,\Xv,\bm{\theta} )} d\Theta_{\text{var}} \nonumber \\
= & \int q_{\bm{\eta}}(\Theta_{\text{var}} )\log \frac{q_{\bm{\eta}}(\Theta_{\text{var}} )p(\yv \vert \Xv,\bm{\theta})}{p(\yv \vert \Xv,\bm{\theta} ,\Theta_{\text{var}} )p(\Theta_{\text{var}} )} d\Theta_{\text{var}} \nonumber \\
= & \int q_{\bm{\eta}}(\Theta_{\text{var}} )\log \frac{q_{\bm{\eta}}(\Theta_{\text{var}} )}{p(\Theta_{\text{var}} )} d\Theta_{\text{var}} -\int q_{\bm{\eta}}(\Theta_{\text{var}} )\log p(\yv \vert \tilde{f}_{\Xv} )d\Theta_{\text{var}} +\log p(\yv\vert \Xv,\bm{\theta}).
\end{align}
Using the fact that $\text{KL}[\cdot \| \cdot] \geq 0$, we have
\begin{align}
\label{eq:variational lower bound}
    \log p(\yv\vert \Xv,\bm{\theta}) & \geq \int q_{\bm{\eta}}(\Theta_{\text{var}} )\log p(\yv \vert \tilde{f}_{\Xv} )d\Theta_{\text{var}} - \text{KL} \left[ q_{\bm{\eta}}(\Theta_{\text{var}} ) \| p(\Theta_{\text{var}}) \right] \nonumber \\
    & = \Eb_{q_{\bm{\eta}}(\Theta_{\text{var}} )} \left[ \log p(\yv \vert \tilde{f}_{\Xv} ) \right] - \text{KL} \left[ q_{\bm{\eta}}(\Theta_{\text{var}} ) \| p(\Theta_{\text{var}}) \right].
\end{align}

\paragraph{Full-training.}
Firstly, we present the joint training of $\bm{\theta}$ and $\bm{\eta}$. The most common approach optimizes the marginal log-likelihood (the left-hand side of \cref{eq:variational lower bound}):
\begin{align}
    \bm{\theta}^{\ast} &=\argmax_{\bm{\theta}} \log p(\yv\vert \Xv,\bm{\theta} ) \\
    &= \argmax_{\bm{\theta}} \log \int p\left( y\vert X,\bm{\theta},\Theta_{\text{var}} \right) p(\Theta_{\text{var}})d\Theta_{\text{var}},
\end{align}
which involves intractable integral in some tasks such as classification. Instead, we optimize the variational lower bound (the right-hand side of \cref{eq:variational lower bound}):
\begin{align}
    \Theta^{\ast} := \argmax_{\bm{\theta},\bm{\eta}} \mathcal{L}(\bm{\theta},\bm{\eta}) =\argmax_{\bm{\theta},\bm{\eta}}\left\{ E_{q_{\bm{\eta}}(\Theta_{\text{var}} )}\left[ \log p(\yv|\tilde{f}_{\Xv} ) \right] -\text{KL} \left[ q_{\bm{\eta}}(\Theta_{\text{var}} )\| p(\Theta_{\text{var}} ) \right] \right\}.
\end{align}

\paragraph{Fine-tuning.}
An alternative training approach is to firstly pre-train the deterministic parameters of feature extractor by standard neural network training, with mean squared error for regression or cross-entropy for classification as the loss function, and then fine-tune the last layer additive GP with fixed features. The objective function is identical to \cref{eq:elbo}, but $\bm{\theta}$ is learned during the pre-training step and is no longer optimized during fine-tuning.

\section{ELBO}
\label{sec:elbo}
\subsection{Assumptions}
Consider the model $y_i = \tilde{f}(\xv_i) + \epsilon_i$ with the i.i.d. noise $\epsilon_i \overset{\text{i.i.d.}}{\sim} \Nc(0, \sigma_{f}^2)$ and $\tilde{f} : \Rb^D \rightarrow \Rb$ is defined in \cref{eq:DAK prediction}. The training dataset is $\mathcal{D} = \{ \Xv, \yv \}$ where $\Xv:=\{ \xv_i \}_{i=1}^N$, $\yv=(y_1,\ldots,y_N)^{\top}$, $\xv_i \in \Rb^D$, $y_i\in\Rb$. $\Theta_{\text{var}}:= \{ \mu ,\{ \zv_{p}\}_{p=1}^{P} \}$ are the variational random variables consisting of Gaussian weights and bias of $P$ units, $\psi$ are the parameters of the NN, $\bm{\sigma}:=(\sigma_1, \ldots, \sigma_p)^{\top}$ are the scale parameters of base GP layers. The variational distributions are $q(\mu)=\Nc(m_{\mu}, \sigma_{\mu}^2)$, $q(\zv_p)=\Nc(\bm{m}_{\zv_p}, \Sv_{\zv_p})$ and the variational priors are $p(\mu)=\Nc(\check{m}_{\mu} ,\check{\sigma}^2_{\mu})$, $p(\zv_p)=\Nc(\check{\bm{m}}_{\zv_p} ,\check{\Sv}_{\zv_p})$. Note that $\Sv_{\zv_p}\in\Rb^{M \times M}$ is a diagonal covariance matrix due to the independence of $\zv_p$, $M$ is the number of inducing points $\Uv$ defined in \cref{eq:GPlayer}, and $\bm{m}_{\zv_p} \in \Rb^M$, $m_{\mu} \in \Rb$, $\sigma_{\mu}^2 \in \Rb$. We derive the ELBO in VI to learn the preditive posterior over the variational variables $\Theta_{\text{var}}:= \{ \mu ,\{ \zv_{p}\}_{p=1}^{P} \}$ parameterized by $\bm{\eta}:=\left\{ \{ \mv_{\zv_{p}},\Sv_{\zv_{p}}\}_{p=1}^{P} , \{m_{\mu},\sigma_{\mu} \} \right\}$, and optimize the deterministic parameters $\bm{\theta}:=\{\psi, \bm{\sigma}\}$.

\subsection{Expected Log Likelihood}
\paragraph{Closed Form}
The \emph{expected log likelihood}, which is the first term in ELBO defined in \cref{eq:elbo}, is given by 
\begin{align}
    {\Eb}_{q_{\bm{\eta}}(\Theta_{\text{var}})} \left[ \log \text{Pr} (\yv \vert \tilde{f}_{\Xv} ) \right]
    &= {\Eb}_{q_{\bm{\eta}}(\Theta_{\text{var}})} \left[ 
    \log \prod_{i=1}^{N} 
    p (y_i \vert \tilde{f}_{\xv_i} )
    \right] \nonumber\\
    &= \sum_{i=1}^{N} 
    {\Eb}_{q_{\bm{\eta}}(\Theta_{\text{var}})} \left[ 
    \log
    p (y_i \vert \tilde{f}_{\xv_i} )
    \right] \nonumber\\
    &= \sum_{i=1}^{N} 
    {\Eb}_{q_{\bm{\eta}}(\Theta_{\text{var}})} \left[ 
    \log
    \Nc( \tilde{f}_i,\hspace{0.2em} \sigma_{f}^2 )
    \right] \nonumber\\
    &= \sum_{i=1}^{N} 
    {\Eb}_{q_{\bm{\eta}}(\Theta_{\text{var}})} \left[ 
    \log \left(
    (2\pi \sigma_{f}^2)^{-\frac{1}{2}}
    \exp\left\{  
        -\frac{ (y_i - \tilde{f}_i)^2 }{2 \sigma_{f}^2}
    \right\}
    \right)
    \right] \nonumber\\
    &= \sum_{i=1}^{N} 
    {\Eb}_{q_{\bm{\eta}}(\Theta_{\text{var}})} \left[
    -\frac{1}{2} \log(2\pi) 
    - \frac{1}{2}\log(\sigma_{f}^2)
    - \frac{1}{2 \sigma_{f}^2}
    (y_i - \tilde{f}_i)^2
    \right] \nonumber\\
    &= - \frac{N}{2} \log(2\pi)
    - \frac{N}{2} \log(\sigma_{f}^2)
    - \frac{1}{2 \sigma_{f}^2}
    \sum_{i=1}^{N}
    {\Eb}_{q_{\bm{\eta}}(\Theta_{\text{var}})} \left[
    (y_i - \tilde{f}_i)^2
    \right] \nonumber\\
    &= - \frac{N}{2} \log(2\pi)
    - \frac{N}{2} \log(\sigma_{f}^2)
    - \frac{1}{2 \sigma_{f}^2}
    \sum_{i=1}^{N} \left(
    \left({\Eb}_{q(\Theta_{\text{var}})} \left[
    (y_i - \tilde{f}_i)
    \right] \right)^2
    + \text{Var}_{q(\Theta_{\text{var}})} \left[
    (y_i - \tilde{f}_i)
    \right]
    \right) \label{eq:evidence halfway},
\end{align}
where
\begin{align}
    \tilde{f}_i
    &= \sum_{p=1}^{P} \sigma_p \Big(
    \begingroup
        \color{blue}
        \underbracket{
            \color{black}
            \phi(h_{\psi}^{[p]}(\xv_i))
        }_{\color{blue}
            :=\bm{\phi}_{i,p}^{\top} \in \Rb^{1 \times M}
        }
    \endgroup
    \zv_p
    \Big)
    + \mu
    \nonumber\\
    &= \sum_{p=1}^{P} \sigma_p \left(
    \bm{\phi}_{i,p}^{\top} \zv_p 
    \right) + \mu.
\end{align}
Recall that the variational assumptions $q(\zv_p)=\Nc(\bm{m}_{\zv_p}, \Sv_{\zv_p})$ and $q(\mu)=\Nc(m_{\mu}, \sigma_{\mu}^2)$, we can infer that
\begin{align}
    \bm{\phi}_{i,p}^{\top} \zv_p + \mu 
    &\sim
    \Nc\left(
    \bm{\phi}_{i,p}^{\top} \bm{m}_{\zv_p} + m_{\mu},\hspace{0.2em}
    \bm{\phi}_{i,p}^{\top} \Sv_{\zv_p} \bm{\phi}_{i,p} + \sigma_{\mu}^2
    \right), \\
    \sigma_p \left(
    \bm{\phi}_{i,p}^{\top} \zv_p 
    \right) + \mu
    & \sim
    \Nc\left(
    \sigma_p ( \bm{\phi}_{i,p}^{\top} \bm{m}_{\zv_p} ) + m_{\mu},\hspace{0.2em}
    \sigma_p^2( \bm{\phi}_{i,p}^{\top} \Sv_{\zv_p} \bm{\phi}_{i,p} ) + \sigma_{\mu}^2
    \right), \\
    \tilde{f}_i = 
    \sum_{p=1}^{P}
    \sigma_p \left(
    \bm{\phi}_{i,p}^{\top} \zv_p 
    \right)+ \mu
    & \sim
    \Nc\left(
    \sum_{p=1}^{P}
    \sigma_p ( \bm{\phi}_{i,p}^{\top} \bm{m}_{\zv_p} )+ m_{\mu},\hspace{0.2em}
    \sum_{p=1}^{P}
    \sigma_p^2( \bm{\phi}_{i,p}^{\top} \Sv_{\zv_p} \bm{\phi}_{i,p} ) + \sigma_{\mu}^2
    \right), \\
    y_i - \tilde{f}_i
    & \sim 
    \Nc\left(
    y_i - 
    \sum_{p=1}^{P}
    \sigma_p ( \bm{\phi}_{i,p}^{\top} \bm{m}_{\zv_p} ) -m_{\mu},\hspace{0.2em}
    \sum_{p=1}^{P}
    \sigma_p^2( \bm{\phi}_{i,p}^{\top} \Sv_{\zv_p} \bm{\phi}_{i,p} ) + \sigma_{\mu}^2
    \right).
\end{align}
Therefore, 
\begin{subequations}\label{eq:exp and var in evidence}
    \begin{align}
        \left({\Eb}_{q(\Theta_{\text{var}})} \left[
        (y_i - \tilde{f}_i)
        \right] \right)^2
        = \left(
         y_i - 
        \sum_{p=1}^{P}
        \sigma_p ( \bm{\phi}_{i,p}^{\top} \bm{m}_{\zv_p} ) -m_{\mu}
        \right)^2,
    \end{align}
    \begin{align}
        \text{Var}_{q(\Theta_{\text{var}})}
        \left[
        (y_i - \tilde{f}_i)
        \right]
        = \sum_{p=1}^{P}
        \sigma_p^2( \bm{\phi}_{i,p}^{\top} \Sv_{\zv_p} \bm{\phi}_{i,p} ) + \sigma_{\mu}^2.
    \end{align}
\end{subequations}
By applying \cref{eq:exp and var in evidence} to \cref{eq:evidence halfway}, we derive the analytical formula for the expected evidence, expressed as
\begin{align}
    {\Eb}_{q_{\bm{\eta}}(\Theta_{\text{var}})} \left[ \log \text{Pr} (\yv \vert \tilde{f}_{\Xv} ) \right]
    &= - \frac{N}{2} \log(2\pi)
    - \frac{N}{2} \log(\sigma_{f}^2) \nonumber\\
    &- \frac{1}{2 \sigma_{f}^2}
    \sum_{i=1}^{N} \left(
        \Big(
         y_i - 
        \sum_{p=1}^{P}
        \sigma_p ( \bm{\phi}_{i,p}^{\top} \bm{m}_{\zv_p} ) -m_{\mu}
        \Big)^2
        + \sum_{p=1}^{P}
        \sigma_p^2( \bm{\phi}_{i,p}^{\top} \Sv_{\zv_p} \bm{\phi}_{i,p} )+ \sigma_{\mu}^2
    \right). \label{eq:evidence final}
\end{align}

\paragraph{Monte Carlo Approximation}
For comparison, we provide the equation for computing the Monte Carlo estimate of the ELBO in the paragraph that follows.
\begin{align}
    {\Eb}_{q_{\bm{\eta}}(\Theta_{\text{var}})} \left[ \log \text{Pr} (\yv \vert \tilde{f}_{\Xv} ) \right]
    &= \sum_{i=1}^{N} 
    {\Eb}_{q_{\bm{\eta}}(\Theta_{\text{var}} )} \left[ 
    \log
    p (y_i \vert \tilde{f}_{\xv_i} )
    \right] \nonumber\\
    & \approx \sum_{i=1}^{N}
    \frac{1}{S}
     \sum_{s=1}^{S}
    \log
    p (y_i \vert \xv_i,\tilde{\Theta}^{(s)}_{\text{var}}, \bm{\theta} ) \nonumber\\
    &= \frac{1}{S} \sum_{i=1}^{N} 
    \sum_{s=1}^{S} 
    \log
    \Nc(y_i \left\vert\right. \tilde{f}_{i}^{(s)},\hspace{0.2em} \sigma_{f}^2 )
    \nonumber\\
    &= \frac{1}{S} \sum_{i=1}^{N} 
    \sum_{s=1}^{S} 
    \log \left(
    (2\pi \sigma_{f}^2)^{-\frac{1}{2}}
    \exp\left\{  
        -\frac{ (y_i - \tilde{f}_{i}^{(s)})^2 }{2 \sigma_{f}^2}
    \right\}
    \right)
    \nonumber\\
    &= \frac{1}{S} \sum_{i=1}^{N} 
    \sum_{s=1}^{S} \left(
    -\frac{1}{2} \log(2\pi) 
    - \frac{1}{2}\log(\sigma_{f}^2)
    - \frac{1}{2 \sigma_{f}^2}
    (y_i - \tilde{f}_{i}^{(s)})^2
    \right) \nonumber\\
    &= - \frac{N}{2} \log(2\pi)
    - \frac{N}{2} \log(\sigma_{f}^2)
    - \frac{1}{2 \sigma_{f}^2}
    \sum_{i=1}^{N}
    \frac{1}{S} \sum_{s=1}^{S}
    (y_i - \tilde{f}_{i}^{(s)})^2, \label{eq:evidence halfway mc approx}
\end{align}
where $S$ is the number of Monte Carlo samples, $\{  \tilde{\mu}^{(s)} ,\{ \tilde{\zv}_{p}^{(s)} \}_{p=1}^{P} \} := \tilde{\Theta}^{(s)}_{\text{var}}$ are the $s$-th Monte Carlo samplings over the variational parameters $\Theta_{\text{var}}$ and $\tilde{\Theta}^{(s)}_{\text{var}} \sim q_{\bm{\eta}}(\Theta_{\text{var}})$, $\tilde{f}_{i}^{(s)}$ is given as follows:
\begin{align}
    \tilde{f}_{i}^{(s)} &:= \tilde{f}(\xv_i;\tilde{\Theta}^{(s)}_{\text{var}},\bm{\theta} ) \nonumber\\
    &= \sum_{p=1}^{P} \sigma_p \Big(
    \begingroup
        \color{blue}
        \underbracket{
            \color{black}
            \phi(h_{\psi}^{[p]}(\xv_i))
        }_{\color{blue}
            :=\bm{\phi}_{i,p}^{\top} \in \Rb^{1 \times M}
        }
    \endgroup
    \tilde{\zv}_p^{(s)} 
    \Big) + \tilde{\mu}^{(s)} \nonumber\\
    &= \sum_{p=1}^{P} \sigma_p \left(
    \bm{\phi}_{i,p}^{\top} \tilde{\zv}_p^{(s)} 
    \right)+ \tilde{\mu}^{(s)}. \label{eq:mc approx mean}
\end{align}
Therefore, we plug \cref{eq:mc approx mean} into \cref{eq:evidence halfway mc approx} and get the the Monte Carlo estimate of the ELBO written in the following formula:
\begin{align}
    {\Eb}_{q_{\bm{\eta}}(\Theta_{\text{var}})} \left[ \log \text{Pr} (\yv \vert \tilde{f}_{\Xv} ) \right]
    &\approx
    - \frac{N}{2} \log(2\pi)
    - \frac{N}{2} \log(\sigma_{f}^2)
    - \frac{1}{2 \sigma_{f}^2}
    \sum_{i=1}^{N}
    \frac{1}{S} \sum_{s=1}^{S}
    \Big(y_i - 
    \sum_{p=1}^{P} \sigma_p \left(
    \bm{\phi}_{i,p}^{\top} \tilde{\zv}_p^{(s)} 
    \Big)- \tilde{\mu}^{(s)}
    \right)^2, \label{eq:evidence final mc approx} \\
    \tilde{\zv}_p^{(s)} &\sim \Nc(\bm{m}_{\zv_p}, \Sv_{\zv_p}),\qquad
    \tilde{\mu}^{(s)} \sim \Nc(m_{\mu}, \sigma_{\mu}^2).
\end{align}

\subsection{KL Divergence}
Since we place Gaussian assumptions over the variational parameters $\Theta_{\text{var}}$,  the \emph{KL divergence}, which is the second term in ELBO defined in \cref{eq:elbo}, is then given by
\begin{align}
    \text{KL} \left[ q(\Theta_{\text{var}} ) \| p(\Theta_{\text{var}}) \right]
    &= \text{KL} \left[ q( \mu ,\{ \zv_{p}\}_{p=1}^{P} ) \Vert p( \mu ,\{ \zv_{p}\}_{p=1}^{P}) \right] \nonumber\\
    & =  
    \text{KL} \left[ q(\mu) \Vert p(\mu) \right] 
    + \sum_{p=1}^{P} 
    \text{KL} \left[ q(\zv_{p}) \Vert p(\zv_{p}) \right],
\end{align}

\begin{align}
     \text{KL} \left[ q(\mu) \Vert p(\mu) \right]
     = \frac{1}{2} \left(
     \frac{\sigma_{\mu}^2}{\check{\sigma}_{\mu}^2} 
     + \frac{(m_{\mu} - \check{m}_{\mu})^2}{\check{\sigma}_{\mu}^2} 
     -\log\left( \frac{\sigma_{\mu}^2}{\check{\sigma}_{\mu}^2} \right)
     -1
     \right),
\end{align}

\begin{align}
    \text{KL} \left[ q(\zv_{p}) \Vert p(\zv_{p}) \right]
    = \frac{1}{2} \sum_{i=1}^{M} \left(
     \frac{[\Sv_{\zv_p}]_{ii}}{[\check{\Sv}_{\zv_p}]_{ii}} 
     + \frac{([\bm{m}_{\zv_p}]_{i} - [\check{\bm{m}}_{\zv_p}]_i)^2}{[\check{\Sv}_{\zv_p}]_{ii}}
     -\log\left( 
     \frac{[\Sv_{\zv_p}]_{ii}}{[\check{\Sv}_{\zv_p}]_{ii}}  
     \right)
     -1
     \right),
\end{align}
where $[\Sv_{\zv_p}]_{ii}$ is the $(i,i)$-th element of the diagonal covariance matrix $\Sv_{\zv_p} \in \Rb^{M \times M}$, $[\bm{m}_{\zv_p}]_{i}$ is the $i$-th element of the mean vector $\bm{m}_{\zv_p} \in \Rb^M$, the approximated posteriors are $q(\mu)=\Nc(m_{\mu}, \sigma_{\mu}^2)$, $q(\zv_p)=\Nc(\bm{m}_{\zv_p}, \Sv_{\zv_p})$ and the priors are $p(\mu)=\Nc(\check{m}_{\mu} ,\check{\sigma}^2_{\mu})$, $p(\zv_p)=\Nc(\check{\bm{m}}_{\zv_p} ,\check{\Sv}_{\zv_p})$.


\subsection{Limitations of the Closed-Form ELBO}

The closed-form ELBO is only applicable to regression problems. In classification, applying the softmax function to $\tilde{f}(\xv;\bm{\theta}, \bm{\eta})$ results in a non-analytic predictive distribution, meaning the ELBO must still be computed via Monte Carlo sampling during training. Similarly, the closed-form expressions for the predictive mean and variance, as provided in \cref{eq:dak inference closed form} in \Cref{sec:uq of inference}, are not applicable to classification but only apply to regression problems.

\section{COMPUTATIONAL COMPLEXITY}
\label{sec:complexity}
In this section, we discuss the computational complexity of various DKL models compared to the proposed DAK method, focusing on the GP layer as the most computationally demanding component. \Cref{tab:complexity supp} shows the computational complexity of our model compared to other state-of-the-art GP and DKL methods.

\begin{table}[ht]
    \caption{Computational complexity of the DKL models for $N$ training points. The reported training complexity is for one iteration. $\hat{M}$ is the number of inducing points in SVGP and KISS-GP, while $M$ is the size of induced grids in DAK, $M < \hat{M}$. $S$ is the number of Monte Carlo samples, $B$ is the size of mini-batch, $D_w$ is the dimension of the NN outputs in DKL, $P$ is the dimension of the outputs after applying linear transformations to the NN outputs in the proposed DAK model. DAK-MC refers to the DAK model using Monte Carlo approximation, while DAK-CF refers to the DAK model using closed-form inference and ELBO.}
    \centering
    \begin{tabular}{lcc}
    \toprule[1pt]
                  & \textbf{Inference}       & \textbf{Training} (per iteration) \\
    \midrule[0.5pt]
    NN + SVGP     & $\Oc(\hat{M}^2 N)$    & $\Oc( S D_w MB + \hat{M}^3)$ \\
    NN + KISS-GP  & $\Oc(D_w \hat{M}^{1+\frac{1}{D_w}})$  & $\Oc(S D_w MB + D_w \hat{M}^{\frac{3}{D_w}})$ \\
    DAK-MC (ours) & $\Oc(SM)$       & $\Oc(SPMB + PM)$   \\
    DAK-CF (ours) & $\Oc(M)$        & $\Oc(PMB + PM)$    \\
    \bottomrule[1pt]
    \end{tabular}
    \label{tab:complexity supp}
\end{table}

\paragraph{Inference Complexity.}
In inference based on induced approximation, computing the multiplication of the inverse of the covariance matrix $k(\Uv, \Uv)$ and a vector takes $\Oc(\hat{M}^2N)$ time for $\hat{M}$ inducing points $\Uv$ and $N$ training points when using SVGP. This cost is reduced by KISS-GP to $\Oc(D \hat{M}^{1+\frac{1}{D}})$ by decomposing the covariance matrix into a Kronecker product of $D$ one-dimensional covariance matrices of the inducing points: $k(\Uv, \Uv) = \bigotimes_{d=1}^{D} k(\Uv^{[d]}, \Uv^{[d]})$. Despite the significant reduction on complexity, it requires inducing points $\Uv$ arranged on a Cartesian grid of size $\hat{M} = \prod_{d=1}^{D} \hat{M}_d$, where $\hat{M}_d$ is the number of inducing points in the $d$-th dimension. In high-dimensional spaces, fixing $\hat{M}$ leads to very small $\hat{M}_d$ per dimension, which can degrade model performance. To address this, we propose the DAK model via sparse finite-rank approximation, which employs an additive Laplace kernel for GPs. The inverse Cholesky factor $\Lv_{\Uv}^{\top}$ for one-dimensional induced grids $\Uv$ of size $M$, where $M < \hat{M}$, as defined in \cref{eq:GPlayer}, is sparse and can be computed in $\Oc(M)$ time.

\paragraph{Training Complexity.}
In training, VI requires computing the ELBO as described in \cref{eq:elbo}, which consists of two terms: the \emph{expected log likelihood} and the \emph{KL divergence} between the variational distributions and priors. 

1) The \emph{expected log likelihood} is usually approximated via Monte Carlo sampling at a cost of $\Oc(S N_{\Theta} N)$, where $S$ is the number of Monte Carlo samples, $N_{\Theta}$ is the total number of variational parameters $\Theta_{\text{var}}$, and $N$ is the number of training points. This complexity can be reduced to $\Oc(S N_{\Theta} B)$ by applying stochastic variational inference with a mini-batch of size $B \ll N$. For DKL models using SVGP and KISS-GP, $\Theta_{\text{var}}$ are inducing variables, and the expectation does not have a closed form, requiring Monte Carlo sampling. In contrast, in the proposed DAK model, $\Theta_{\text{var}}= \{ \{ \zv_{p}\}_{p=1}^{P}, \mu \}$ consists of independent Gaussian weights $\zv_p\in \Rb^M$ and bias $\mu$. This allows us to derive an analytical form for this term, as shown in \cref{eq:evidence final} in \Cref{sec:elbo}, reducing the computational cost to $\Oc(N_{\Theta} B) = \Oc(PM B)$ when using a mini-batch of size $B$.

2) The \emph{KL divergence} between two Gaussian distributions can be computed in closed form. This leads to a linear time complexity of $\Oc(N_{\Theta})$ if the parameters $\Theta_{\text{var}}$ are independent, or cubic time $\Oc(N_{\Theta}^3)$ if they are fully correlated. In SVGP and KISS-GP, $\Theta_{\text{var}}$ represents fully correlated Gaussian distributed inducing variables, so computing the KL divergence takes $\Oc(\hat{M}^3)$ for SVGP. In KISS-GP, this can be reduced to $\Oc(D \hat{M}^{\frac{3}{D}})$ using fast eigendecomposition of Kronecker matrices. In the DAK model, the weights $\{\zv_p\}_{p=1}^{P}$ as defined in \cref{eq:GPlayer} are independent Gaussian random variables, allowing the KL divergence to be computed in $\Oc(N_{\Theta}) = \Oc(PM)$ time, where $P$ is the number of base GP layers.

\section{ADDITIONAL DISCUSSIONS}

Although interpretability is one advantage of additive models, the main motivation for replacing a GP layer with an additive GP layer in our work is to handle high-dimensional data. When the input dimension is low, it is reasonable that GPs are superior to additive GPs since the additive kernel is an approximated and restrictive kernel. However, when the input dimension increases, the computational complexity grows considerably even in GPs with sparse approximation. For example, in DKL, the output dimension of NN encoder is usually chosen as small as 2, while in pixel data experiments, DKL cannot handle the computation associated with the dimensionality when the output dimension of ResNet is 512 or more. Although DKL is superior in low-dimensional and simple cases, we view additive structure as a necessary component to achieve scalability and good performance with high-dimensional data.

\subsection{Why choosing the induced grids instead of learning the inducing points?}

From an approximation accuracy point of view, there are two separate strategies to increase the accuracy. The first one is to learn the inducing point locations. The second one, however, is to simply increase the number of inducing points on a pre-specified finer grid. The second method is much easier to implement and has a theoretical guarantee by the GP regression theory: as the inducing points become dense in the input region, the approximation will become exact. In contrast, the first approach does not have such a favorable theoretical guarantee. 

The second approach would become difficult to use for many existing methodologies as in general the computational cost would scale as $\mathcal{O}(M^3)$ with $M$ inducing points, which is particularly problematic in high dimensions. 
This difficulty is resolved by additive GPs, since approximating an additive GP boils down to approximating one dimensional GPs, which can be accomplished by using a set of pre-specified inducing points on a fine grid in 1-D. One major benefit of the proposed methodology is that the computation now scales at $\mathcal{O}(M)$, enabled by the Markov kernel and the additive kernel. Therefore, a large number of inducing points can be used in an efficient way. 

The proposed method also has several additional benefits: 1) It can decouple to some extent the neural network component and GP component by avoiding learning the inducing points, which may help reduce overfitting/overconfidence; 2) The equivalence to BNN holds exactly with the fixed inducing points, whereas for learned inducing points, this BNN equivalence breaks down, and the proposed computation/training framework would not be possible to carry through; 3) It can simplify the overall optimization since there is no need to learn the inducing points.

\subsection{Limitations and future directions}

Generally, a finer grid will lead to better approximations, but the number of parameters to be trained will also increase. Therefore, there is a trade-off between the accuracy and the computational cost that we can afford. This current work is using a specific Laplace kernel, which can utilize sparse Cholesky decomposition. More general kernels may result in more computational complexity but better representation power of the model. In addition, the current variational family is restricted under mean-field assumptions. A more general variational family, e.g. full/low-rank covariance, may lead to superior performance in some applications.

\section{EXPERIMENTAL DETAILS}
\label{sec:expdetail}
In this section, we provide additional details regarding the experiments.

\subsection{Benchmarks for Regression}
\label{subsec:regression supp}
\paragraph{Experiment Setup}
For all models, the NN architecture is a fully connected NN with rectified linear unit (ReLU) activation function \citep{nair2010rectified} and two hidden layers containing 64 and 32 neurons, respectively, structured as $D \rightarrow 64 \rightarrow 32 \rightarrow D_w$, where $D$ is the input feature size (also the size of input $\Xv$) and $D_w$ is the output feature size. The models are evaluated with $D_w=16$, 64, and 256, respectively. The number of Monte Carlo samples is set to 8 during training and 20 during inference.

The NN is a deterministic model, and we use the negative Gaussian log-likelihood as the loss function to quantify the uncertainty of the NN outputs and compute the NLPD.

For NN+SVGP, the inducing points are set to the size of 64 in $D_w$ dimension. We implement the \texttt{ApproximateGP} model in GPyTorch \citep{gardner2018gpytorch}, defining the inducing variables as variational parameters, and use \texttt{VariationalELBO} in GPyTorch to perform variational inference and compute the loss.

SV-DKL is originally designed for classification, so for a fair comparison in regression tasks, we modify it by first applying a linear embedding layer $\Wv: \Rb^{D_w} \rightarrow \Rb^P$ with $P=16$ and normalizing the outputs to the interval $[0,1]$ for each base GP, similar to the DAK model. To adapt the additive GP layer for regression, we remove the softmax function from the model in eq. (1) of \citep{wilson2016stochastic}. Given training data $\{ \xv_i, \yv_i \}_{i=1}^{N}$, the model is modified as follows:
\begin{align}
    p(\yv_i \vert \fv_i, A) = \mathcal{A}(\fv_i)^{\top} \yv_i
\end{align}
where $\fv_i \in \Rb^P$ is a vector of independent GPs followed by a linear mixing layer $\mathcal{A}(\fv_i) = A \fv_i$, with $A \in \Rb^{C \times P}$ as the transformation matrix. Here, $C=1$ for single-task regression. For each $p$-th GP ($1 \leq p \leq P$) in the additive GP layer, the corresponding inducing variables $\uv_p$ are set to the size of 64 and treated as variational parameters for training. We use the \texttt{GridInterpolationVariationalStrategy} model with \texttt{LMCVariationalStrategy} in GPyTorch to perform KISS-GP with variational inducing variables, augmented by a linear mixing layer.

For AV-DKL, the inducing points are set to size of 64 in $D_{w}$ dimension. We implement the AV-DKL model based on the source code~\cite{matias2024amortized}.

Both DAK-MC and DAK-CF use the same additive GP layer size as SV-DKL, with $P=16$, and employ fixed induced grids $\Uv = \{1/8, 2/8, \ldots, 7/8\}$ of size 7 for each base GP, which is much smaller than that of SV-DKL.

\paragraph{Metrics}
Let $\{\xv_t, y_t\}_{t=1}^{T}$ represent a test dataset of size $T$, where $\mu_t$ and $\sigma_t^2$ are the predictive mean and variance. We evaluate model performance using two common metrics: Root Mean Squared Error (RMSE) and Negative Log Predictive Density (NLPD).

RMSE is widely used to assess the accuracy of predictions, measuring how far predictions deviate from the true target values. It is calculated as:
\begin{align}
    \text{RMSE} = \sqrt{ \frac{1}{T} \sum_{t=1}^{T}(y_t - \mu_t)^2 }.
\end{align}

NLPD is a standard probabilistic metric for evaluating the quality of a model's uncertainty quantification. It represents the negative log likelihood of the test data given the predictive distribution. For GPs, NLPD is calculated as:
\begin{align}
    \text{NLPD}
    &= - \sum_{t=1}^{T} \log p(y_t = \mu_t \vert \xv_t) \\
    &= \frac{1}{T}
    \sum_{t=1}^{T} \Big[
    \frac{(y_t - \mu_t)^2}{2\sigma_t^2} + \frac{1}{2} \log(2\pi \sigma_t^2)
    \Big].
\end{align}
Both RMSE and NLPD are widely used in the GP regression literature, where smaller values indicate better model performance.

\paragraph{Computing Infrastructure}
The experiments for regression were run on Macbook Pro M1 with 8 cores and 16GB RAM.

\subsection{Benchmarks for Classification}
\label{subsec:classification supp}
We use PyTorch \citep{paszke2019pytorch} baseline of NN models, GPyTorch \citep{gardner2018gpytorch} baseline of SVGP and SV-DKL models. In classification tasks, we apply a softmax likelihood to normalize the output digits to probability distributions. The NN is a deterministic model trained via negative log-likelihood loss, while DKL and DAK models are trained via ELBO loss. The setting of all training tasks are described in \Cref{tab:model classification} and \Cref{tab:optimizer classification}.

SVGP is originally designed for single-output regression. To make it fit for multi-output classification, we used \texttt{IndependentMultitaskVariationalStrategy} in GPyTorch to implement the multi-task \texttt{ApproximateGP} model, and use \texttt{VariationalELBO} with \texttt{SoftmaxLikelihood} in GPyTorch to perform variational inference and compute the loss. 

For SV-DKL, we employed the same \texttt{VariationalELBO} with \texttt{SoftmaxLikelihood} as the variational loss objective. \texttt{GridInterpolationVariationalStrategy} is applied within \texttt{IndependentMultitaskVariationalStrategy} to perform additive KISS-GP approximation. For each KISS-GP unit, we used $64$ variational inducing points initialized on a grid of size $[-1,1]$. 

For DAK, we implemented DAK-MC using Monte Carlo estimation given the intractable softmax likelihood. We employed fixed induced grids $\Uv=\{ -31/32, -30/32, \ldots, 30/32, 31/32 \}$ of size 63 for each base GP component.

\begin{table}[ht]
\caption{Model architectures for image classification on MNIST, CIFAR-10 and CIFAR-100.}
\centering
\resizebox{0.7\linewidth}{!}{
\begin{tabular}{l|l|ccc}
\toprule[1pt]
Model                   & Hyper-parameter          & MNIST       & CIFAR-10    & CIFAR-100   \\
\midrule[0.5pt]
\multirow{4}{*}{NN+SVGP}   & Feature extractor        & CNN         & ResNet-18   & ResNet-34   \\
                        & NN out features $D_w$         & 128         & 512         & 512         \\
                        & Embedding features $P$               & 16          & 64          & 128         \\
                        & \# inducing points $\hat{M}$      & 512         & 512         & 512         \\
                        & \# epochs       & 20         & 200         & 200         \\
                        & Training strategy      & Full-training         & Full-training         & Fine-tuning         \\
\midrule[0.5pt]
\multirow{5}{*}{SV-DKL} & Feature extractor        & CNN         & ResNet-18   & ResNet-34   \\
                        & NN out features $D_w$         & 128         & 512         & 512         \\
                        & Embedding features $P$               & 16          & 64          & 128         \\
                        & \# inducing points $\hat{M}$      & 64          & 64          & 64          \\
                        & Grid bounds              & {[}-1,1{]} & {[}-1,1{]} & {[}-1,1{]} \\
                        & \# epochs       & 20         & 200         & 200         \\
                        & Training strategy       & Full-training         & Full-training         & Fine-tuning         \\
\midrule[0.5pt]
\multirow{4}{*}{DAK}    & Feature extractor        & CNN         & ResNet-18   & ResNet-34   \\
                        & NN out features $D_w$         & 128         & 512         & 512         \\
                        & Embedding features $P$               & 16          & 64          & 128         \\
                        & \# induced interpolation $M$ & 63          & 63          & 63         \\
                        & \# epochs       & 20         & 200         & 200         \\
                        & Training strategy      & Full-training         & Full-training         & Full-training         \\
\bottomrule[1pt]
\end{tabular}

}
\label{tab:model classification}
\end{table}

\paragraph{MNIST} We used a CNN implemented in PyTorch as the feature extractor: \texttt{Conv2d}(1,32,3) $\rightarrow$ \texttt{Conv2d}(32,64,3) $\rightarrow$ \texttt{MaxPool2d}(2) $\rightarrow$ \texttt{Dropout}(0.25) $\rightarrow$ \texttt{Linear}(9216,128) $\rightarrow$ \texttt{Dropout}(0.5). To make a fair comparison, for both SV-DKL and DAK, we applied an embedding module through a linear layer that transform $128$ output features into $P=16$ base GP channels. 

\paragraph{CIFAR-10} We used a ResNet-18 as the feature extractor followed by a linear embedding layer that compressed the $512$ output features into $P=64$ base GP channels. 

\paragraph{CIFAR-100} We used a pretrained ResNet-34 as the feature extractor for SV-DKL and fine-tuned GP output layers since SV-DKL struggled to fit using full-training. For proposed DAK, we used full-training. The number of base GP channels is selected as $P=128$. 

\begin{table}[ht]
\caption{Details of training optimizer for image classification on MNIST, CIFAR-10 and CIFAR-100.}
\centering
\resizebox{0.7\linewidth}{!}{

\begin{tabular}{l|ccc}
\toprule[1pt]
Optimization      & MNIST                                                             & CIFAR-10                                                                                                  & CIFAR-100                                                                                                 \\
\midrule[0.5pt]
Optimizer         & Adadelta                                                          & SGD                                                                                                       & SGD                                                                                                       \\
Initial lr.       & 1.0                                                               & 0.1                                                                                                       & 0.1                                                                                                       \\
Weight decay      & 0.0001                                                            & 0.0001                                                                                                    & 0.0001                                                                                                    \\
Scheduler         & StepLR                                                            & CosineAnnealingLR                                                                                         & CosineAnnealingLR                                                                                         \\
\midrule[0.5pt]
Data Augmentation & MNIST                                                             & CIFAR-10                                                                                                  & CIFAR-100                                                                                                 \\
\midrule[0.5pt]
RandomCrop        & -                                                                 & size=32, padding=4                                                                                        & size=32, padding=4                                                                                        \\
HorizontalFlip    & -                                                                 & p=0.5                                                                                                     & p=0.5                                                                                                     \\
\bottomrule[1pt]
\end{tabular}
}
\label{tab:optimizer classification}
\end{table}

\paragraph{Additional Benchmark.}  \citet{matias2024amortized} proposed Amortized Variational DKL (AV-DKL), which is a variant SV-DKL using amortization network to compute the inducing locations and variational parameters, thus attenuating the overcorrelation of NN extracted features. AV-DKL is included as the additional benchmark for classification tasks in \Cref{tab:img avdkl}. The training recipe is the same with SV-DKL.

\begin{table*}[ht]
\caption{\small{Accuracy, NLL, ECE for AV-DKL, SV-DKL, DAK-MC on CIFAR-10/100 averaged over 3 runs. CIFAR-10 uses ResNet-18 with 64 features extracted; CIFAR-100 uses ResNet-34 with 512 features. The best results are highlighted in \textbf{bold}; the second best results are highlighted by \underline{underline}.}}
\centering
\vspace{-0.1cm}
\resizebox{\linewidth}{!}{%
\begin{tabular}{rccclccc}
\toprule[1pt]
\multicolumn{1}{l}{} & \multicolumn{3}{c}{Batch size: 128}  &  & \multicolumn{3}{c}{Batch size: 1024} \\ \cline{2-4} \cline{6-8} \vspace{-8pt} \\
\multicolumn{1}{l}{} & AV-DKL & SV-DKL & \cellcolor{Gray} DAK-MC &   & AV-DKL  & SV-DKL & \cellcolor{Gray} DAK-MC \\ 
\midrule[1pt]
CIFAR-10 - Acc. (\%) $\uparrow$    & \underline{94.23 $\pm$ 0.65}  & 93.44 $\pm$ 0.28    &  \cellcolor{Gray} \textbf{94.81 $\pm$ 0.13}   &     &  \textbf{93.32} $\pm$ \textbf{0.13}        & 90.22 $\pm$ 1.42       & \cellcolor{Gray} \underline{93.02 $\pm$ 0.18}        \\
NLL $\downarrow$     & 0.352 $\pm$ 0.084    & \underline{0.312 $\pm$ 0.033}       &  \cellcolor{Gray} \textbf{0.256} $\pm$ \textbf{0.014}     &      & \underline{0.439 $\pm$ 0.022}         & 0.485 $\pm$ 0.061       & \cellcolor{Gray} \textbf{0.345 $\pm$ 0.001}    \\
ECE $\downarrow$      & 0.048 $\pm$ 0.006    & \underline{0.046 $\pm$ 0.003}       &  \cellcolor{Gray} \textbf{0.039 $\pm$ 0.002}          &     & \underline{0.054 $\pm$ 0.001}       & 0.060 $\pm$ 0.004       & \cellcolor{Gray} \textbf{0.052 $\pm$ 0.001}           \\
\midrule[1pt]
CIFAR-100 -  Acc. (\%) $\uparrow$    & \textbf{77.47 $\pm$ 0.19}  & 74.52 $\pm$ 0.13       & \cellcolor{Gray}  \underline{76.75 $\pm$ 0.18}     &     &  \textbf{77.07 $\pm$ 0.10}        & 66.54 $\pm$ 0.74       & \cellcolor{Gray} \underline{70.38 $\pm$ 1.25}        \\
NLL $\downarrow$     & 1.787 $\pm$ 0.011    & \underline{1.041 $\pm$ 0.007}       & \cellcolor{Gray}  \textbf{1.001 $\pm$ 0.027}     &      & 2.326 $\pm$ 0.030    & \underline{1.738 $\pm$  0.058}      & \cellcolor{Gray} \textbf{1.203 $\pm$ 0.040}        \\
ECE $\downarrow$      & 0.166 $\pm$ 0.002    & \underline{0.049 $\pm$ 0.002}       & \cellcolor{Gray}  \textbf{0.041 $\pm$ 0.004}        &     & 0.175 $\pm$ 0.001         & \underline{0.148 $\pm$ 0.007}       &\cellcolor{Gray}  \textbf{0.056 $\pm$ 0.006}           \\
\bottomrule[1pt]
\end{tabular}
}
\vspace{-0.2cm}
\label{tab:img avdkl}
\end{table*}

\paragraph{Metrics} 
We evaluate model performance using four common metrics: Top-1 accuracy, ELBO, Negative Log Likelihood (NLL), and Expected Calibration Error (ECE). 

ECE is a metric used to quantify the degree of ``calibration'' of a probabilistic model in UQ, specifically for classification problems. It is defined as the weighted average of the absolute difference between the model's predicted probability (confidence) and the actual outcome (accuracy) over several bins of predicted probability. Mathematically, ECE is given by:
\begin{align}
    \text{ECE} =\sum_{m=1}^{M} \frac{\left| B_{m} \right|}{n} \left| \text{acc} (B_{m})-\text{conf} (B_{m}) \right|,
\end{align}
where $M$ is the number of bins into which the confidence values are partitioned, $B_m$ is the set of indices of samples whose predicted confidence falls into the $m$-th bin, $n$ is the total number of samples.

\paragraph{Computing Infrastructure}
The experiments for classification were run on a Linux machine with NVIDIA RTX4080 GPU, and 32GB of RAM.

\subsection{Additional Tables and Figures}
\label{sec:additional exp results}

\begin{figure}[ht]
\centering
\subfloat[$\begin{gathered}\text{DKL: last-layer lr} =0.01.\\ \text{Training loss:} -0.21.\end{gathered}$]{\includegraphics[width=.3\textwidth]{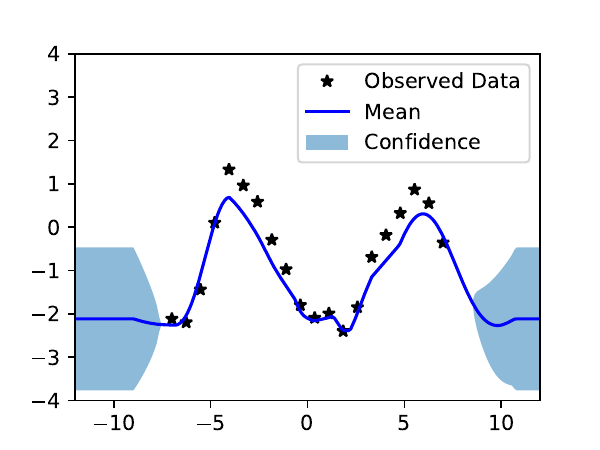}}
\subfloat[$\begin{gathered}\text{DKL: last-layer lr} =0.001.\\ \text{Training loss: } -0.07.\end{gathered}$]{\includegraphics[width=.3\textwidth]{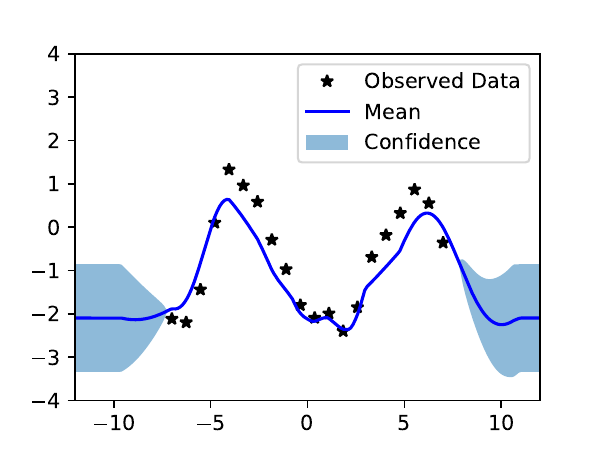}}
\subfloat[$\begin{gathered}\text{DKL: last-layer lr} =0.0001.\\ \text{Training loss: } 0.22.\end{gathered}$]{\includegraphics[width=.3\textwidth]{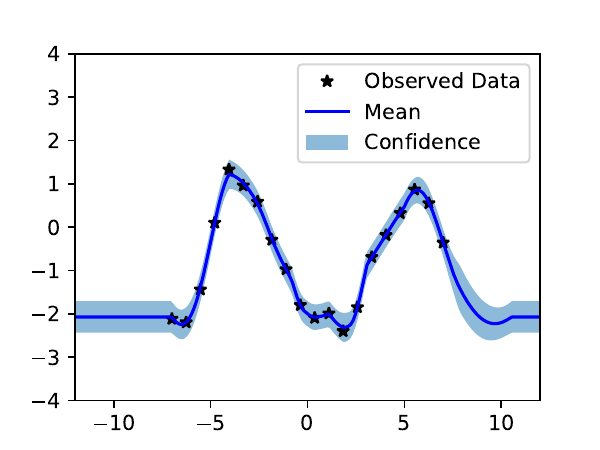}}

\subfloat[$\begin{gathered}\text{DAK: last-layer lr} =0.1.\\ \text{Training loss: } 0.10.\end{gathered}$]{\includegraphics[width=.3\textwidth]{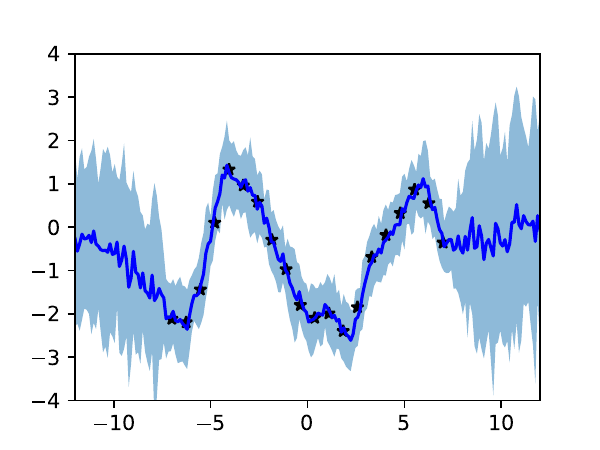}}
\subfloat[$\begin{gathered}\text{DAK: last-layer lr} =0.01.\\ \text{Training loss: } 0.10.\end{gathered}$]{\includegraphics[width=.3\textwidth]{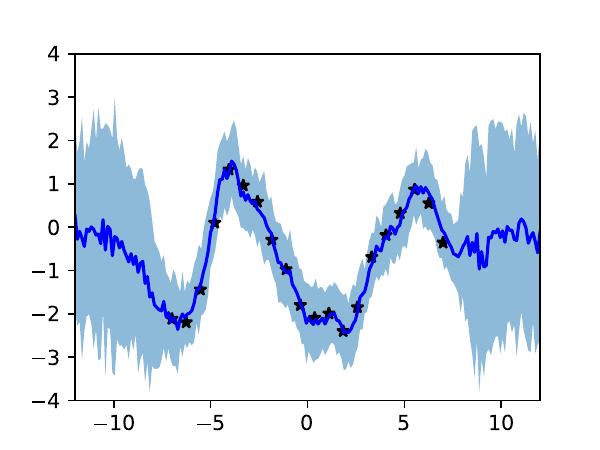}}
\subfloat[$\begin{gathered}\text{DAK: last-layer lr} =0.001.\\ \text{Training loss: } 0.22.\end{gathered}$]{\includegraphics[width=.3\textwidth]{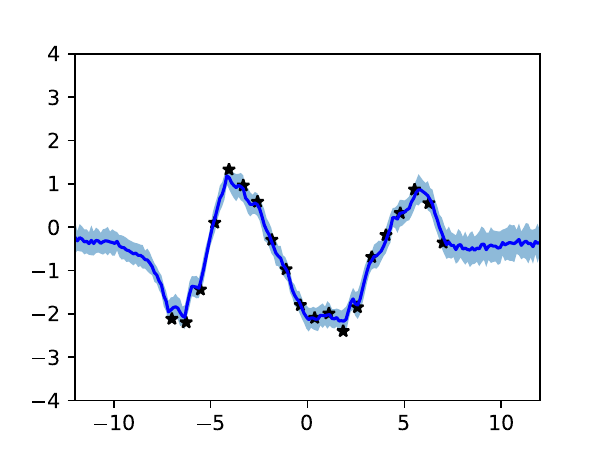}}

\caption{Results on 1D regression with different last-layer learning rates. The learning rate of NN feature extractor is set as $0.01$. (a)--(f) shows the regression fits and corresponding training losses. DAK fits for the same learning rate strategy with NN feature extractor (lr=0.01), while DKL requires a separate tuning for last-layer learning rate of GPs. Additionally, a better training loss does not necessarily prevent overfitting for DKL.}
\label{fig:dkl lr}
\end{figure}

\paragraph{Choices of learning rates.}
We evaluate the choices of learning rates on 1D regression examples. DKL requires a separate tuning of the learning rate of the GP covariance parameters, which differs from the learning rate of the NN feature extractor. In \Cref{fig:dkl lr}, we choose the learning rate of the NN feature extractor as $0.01$, while the learning rate of the GP covariance is set to different values. (a)-(c) show that different learning rates of covariance in DKL result in different predictive posterior. In particular, although the training losses for DKL in both (a) and (b) are minimal, the regressions do not fit well. On the other hand, DAK does not need a distinct recipe for tuning GP covariances because of the BNN interpretation. Furthermore, the poor posterior is indicated by the higher training loss, as illustrated in (d)-(f).

\begin{figure}[ht]
\centering
\subfloat[Test Error (\%).]{\includegraphics[width=.3\textwidth]{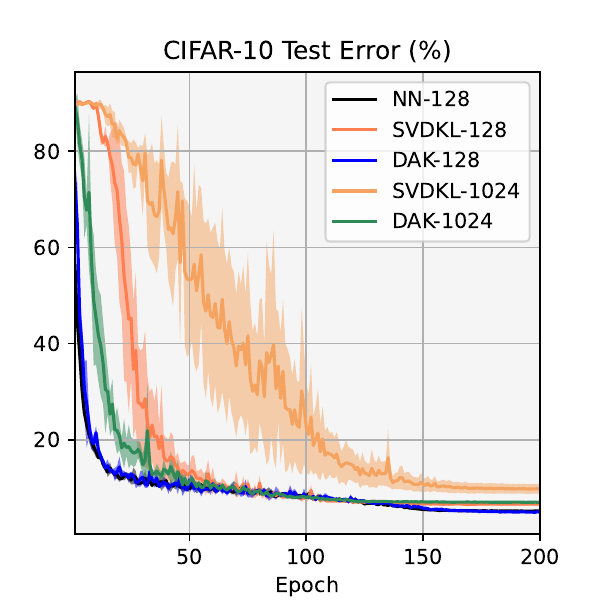}}
\subfloat[Test NLL.]{\includegraphics[width=.3\textwidth]{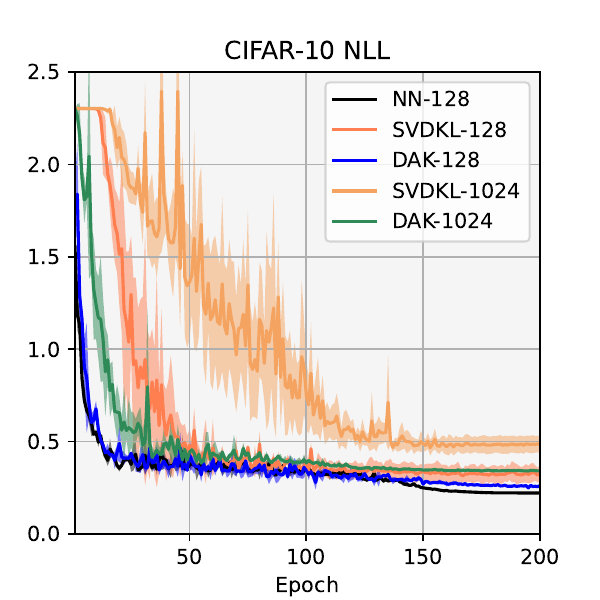}}
\subfloat[ELBO.]{\includegraphics[width=.3\textwidth]{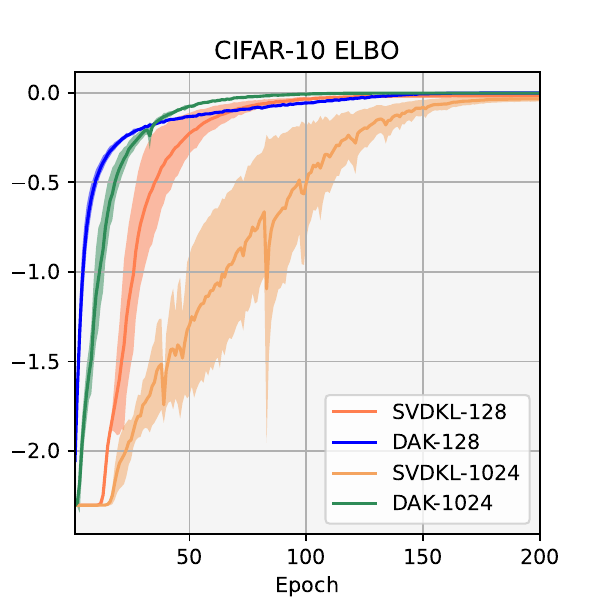}}
\caption{Test errors, test NLLs, ELBOs of NN, SVDKL, and DAK curves with batch size of 128/1024 for CIFAR-10 averaged on 3 runs. DAK outperforms SVDKL on both test error and NLL along the training epochs. Additionally, SVDKL degrades more and struggles to fit when the batch size becomes larger.}
\label{fig:cifar10 curves}
\end{figure}

\begin{figure}[ht]
\centering
\subfloat[Test Error (\%).]{\includegraphics[width=.3\textwidth]{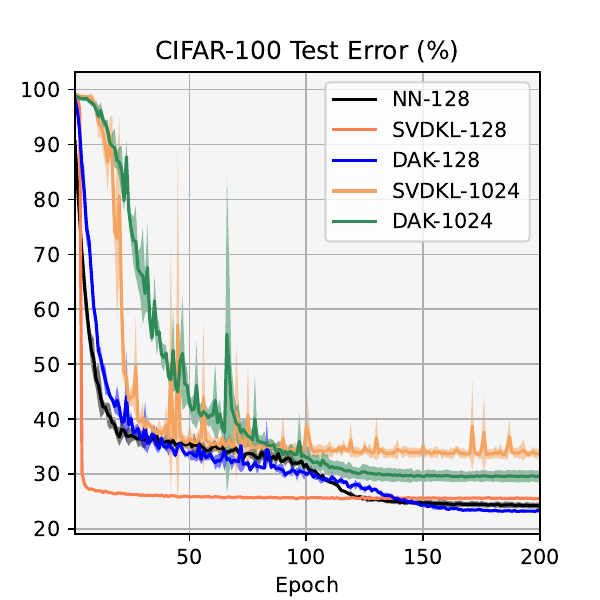}}
\subfloat[Test NLL.]{\includegraphics[width=.3\textwidth]{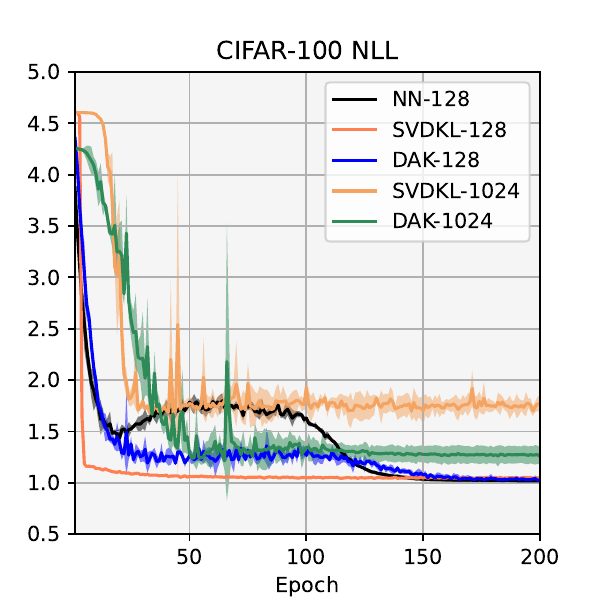}}
\subfloat[ELBO.]{\includegraphics[width=.3\textwidth]{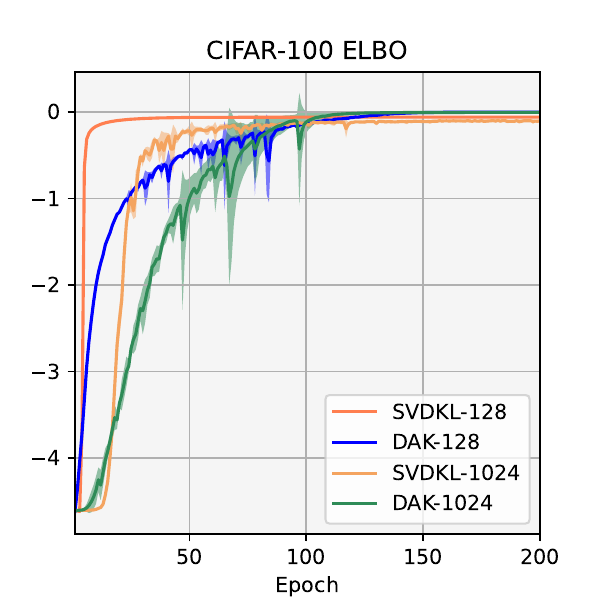}}
\caption{Test errors, test NLLs, ELBOs of NN, SVDKL, and DAK curves with batch size of 128/1024 for CIFAR-100 averaged on 3 runs. DAK trained NN and last-layer additive GPs jointly, while SVDKL used the pre-trained NN and fine-tuned the last-layer GP since SVDKL struggles to fit using full-training. DAK outperforms SVDKL on both test error and NLL along the training epochs. SVDKL struggled to fit in high-dimensional multitask cases, indicating the necessity of pre-training in SVDKL. However, DAK fitted well with high dimensionality and large batch sizes.}
\label{fig:cifar100 curves}
\end{figure}

\paragraph{Learning curves.} We plot the learning curves of CIFAR-10/100 in \Cref{fig:cifar10 curves} and \ref{fig:cifar100 curves}. The learning curves of SVDKL in \Cref{fig:cifar10 curves} is more unstable, with many significant spikes, and the convergence is slower than DAK. Futhermore, SVDKL struggles to fit with full-training in CIFAR-100, and a pretrained feature extractor is used in CIFAR-100. Therefore, the learning curves of SVDKL look smoothing, but DAK fits well with full-training in CIFAR-100.

\end{document}